%% file: main.tex
\documentclass[10pt,twocolumn,letterpaper]{article}

\usepackage{iccv}              

\usepackage{multirow}
\usepackage{pifont}
\usepackage{stfloats}
\input{preamble}

\def\paperID{} 
\def\confName{ICCV}
\def\confYear{2025}

\title{\model\icon: Instructional Editing and Reasoning \\ Video Concepts with Grounded Generation}

\author{%
Shoubin Yu$^{1,3}$\thanks{Equal contribution.}\thanks{Work done during internship at Adobe Research.} \quad 
Difan Liu$^{1 *}$ \quad 
Ziqiao Ma$^{1,2 *}$ \quad \\
Yicong Hong$^{1}$ \quad  
Yang Zhou$^{1}$ \quad  
Hao Tan$^{1}$ \quad 
Joyce Chai$^{2}$ \quad 
Mohit Bansal$^{3}$ \\  \\
$^{1}$Adobe Research \quad
$^{2}$University of Michigan \quad
$^{3}$UNC Chapel Hill 
\\
{{ \tt \normalsize \href{https://veggie-gen.github.io/}{\textcolor{magenta}{https://veggie-gen.github.io/}} }}}

\begin{document}

\input{teaser}
\maketitle

\input{sec/0_abstract}  
\input{sec/1_intro}

\input{sec/2_related_work}

\input{sec/3_method}
\input{sec/4_exp}

\input{sec/5_conclusion}
{
    \small
    \bibliographystyle{ieeenat_fullname}
    \bibliography{main}
}

\input{sec/X_suppl}

\end{document}

%% file: preamble.tex
\usepackage{multirow}
\usepackage{pifont}

\usepackage{soul}
\usepackage{xcolor}
\usepackage{color, colortbl}
\usepackage{tabu}
\usepackage{makecell}
\usepackage{threeparttable}
\usepackage{booktabs}
\usepackage[utf8]{inputenc} 
\usepackage[T1]{fontenc}    
\usepackage{amsfonts}       
\usepackage{nicefrac}       
\usepackage{microtype}      
\usepackage{algpseudocode}
\usepackage{times}
\usepackage{bbding}
\usepackage{tabularx}
\usepackage{caption}

\usepackage{wrapfig}
\usepackage{subfloat}
\usepackage{floatrow}
\usepackage{algorithm}

\usepackage{xspace}
\usepackage{appendix}
\usepackage{array,etoolbox}
\usepackage{tcolorbox}

\definecolor{azureblue}{rgb}{0,0.5,1}
\definecolor{color1}{HTML}{006EB8}
\definecolor{color2}{HTML}{009B55}

\newcommand{\cmark}{\ding{51}}

\usepackage[pagebackref,breaklinks,colorlinks,allcolors=iccvblue]{hyperref}
\usepackage[capitalize]{cleveref}
\crefname{section}{Sec.}{Secs.}
\Crefname{section}{Section}{Sections}
\Crefname{table}{Table}{Tables}
\crefname{table}{Tab.}{Tabs.}
\Crefname{figure}{Table}{Tables}
\crefname{figure}{Fig.}{Figs.}
\crefname{appendix}{Sec.}{Secs.}
\Crefname{appendix}{Section}{Sections}

\definecolor{darkgreen}{rgb}{0,0.5,0}

\definecolor{azureblue}{rgb}{0,0.5,1}

\newcommand{\model}{\texttt{VEGGIE}\xspace}
\newcommand{\benchmark}{\texttt{VEG-Bench}\xspace}
\newcommand{\dataset}{\texttt{VEG-Edit}\xspace}
\newcommand{\icon}{\includegraphics[height=0.9em]{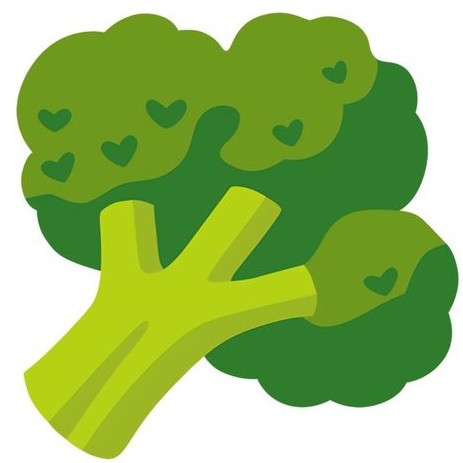}}

\definecolor{gg}{HTML}{F2DFC2}
\definecolor{ggg}{gray}{0.65}
\newcolumntype{a}{>{\columncolor{gg}}c}

\definecolor{iccvblue}{rgb}{0.21,0.49,0.74}

%% file: sec/0_abstract.tex
\begin{abstract}
While recent video diffusion models enable video editing, unifying diverse instructional editing tasks (e.g., add, remove, modify) under a single framework remains a significant challenge.
In this paper, we introduce \model, a Video Editor with Grounded Generation from Instructions, a simple end-to-end framework that unifies video concept editing, grounding, and reasoning based on diverse user instructions. 
Specifically, given a video and text query, \model first utilizes an MLLM to interpret user intentions in instructions and ground them to the video contexts, generating frame-specific grounded task queries for pixel-space responses. 
A diffusion model then renders these plans and generates edited videos that align with user intent.
To support diverse tasks and complex instructions, we employ a curriculum learning strategy: first aligning the MLLM and video diffusion model with large-scale instructional image editing data, followed by end-to-end fine-tuning on high-quality multitask video data. 
Additionally, we introduce a novel data synthesis pipeline to generate paired instructional video editing data for model training. 
It transforms static image data into diverse, high-quality video editing samples by leveraging Image-to-Video models to inject dynamics.
\model shows strong performance in instructional video editing with different editing skills, outperforming the best instructional baseline as a versatile
model, while other models struggle with multi-tasking.
\model also excels in video object grounding and reasoning segmentation, where other baselines fail. 
We further reveal how the multiple tasks help each other and highlight promising applications like zero-shot multimodal instructional and in-context video editing.
\end{abstract}

%% file: sec/1_intro.tex
\vspace{-20pt}
\section{Introduction}
\label{sec:intro}

Building on the advances in Video Diffusion Models (VidDMs)~\cite{ho2022vdm,ho2022imagenv,blattmann2023svd,brooks2024video,yang2024cogvideox}, video editing methods have emerged as video design tools, allowing users to manipulate video concepts such as adding, removing, altering objects and style translation~\citep{geyer2023tokenflow,qi2023fatezero,wang2024videocomposer,yu2023inpaint,zhang2023avid}.
To enhance user experiences, instructional video editing methods~\cite{qin2024instructvid2vid,zhang2024effived} have been developed, using triples of text prompts, source videos, and target videos for training. 
Due to their limited performance in understanding user intent and multimodal semantics~\cite{fu2023guiding}, several methods have incorporated multimodal large language models (MLLMs) to handle complex instructions/reasoning~\cite{huang2024smartedit,yoon2024raccoon,fu2023guiding}.

However, existing methods fall short of the goal of a simple, versatile video concept editor, facing three primary challenges. 
First, most methods are not end-to-end, requiring intermediate layout/mask/human or model caption guidance~\cite{yoon2024raccoon,qi2023fatezero,wang2024genartist,ku2024anyv2v}, which adds workload on users and disrupts a seamless editing experience.
Second, existing pipelines connecting MLLMs to VidDMs require multiple training objectives beyond simple pixel-space diffusion loss, such as language loss~\cite{yoon2024raccoon} or mask losses~\cite{wu2024towards}. 
This increases optimization difficulty and often requires additional hyperparameter tuning or annotations. 
Third, existing video editing models, both instructional and non-instructional, struggle with handling other diverse editing tasks, ranging from addition, and deletion to stylization. 
For example, LGVI~\cite{wu2024towards} fails in global edits such as stylization and color change, while VidToMe~\cite{li2024vidtome} struggles with local edits such as adding or removing objects. 
These methods also struggle with input videos that contain multiple objects or when user instructions require complex reasoning.

These challenges result from two limitations:
First, there is a lack of multitasking fine-tuning on well-curated instructional video editing datasets that span a broad range of skills.
Second, models often lack two critical capabilities needed to interpret user intentions and accurately locate concepts: \textit{multimodal reasoning} to infer the intended modification from the user’s instruction; and \textit{grounding} language to the input video to precisely identify the region or object to be edited.
For example, in Figure~\ref{fig:teaser}, one can effortlessly locate the girl given ``\textit{identify the little girl}.'' 
When asked to ``\textit{add a hat to the little girl},'' we intuitively imagine the hat placed on her head from commonsense, even without seeing an actual hat.

To address these challenges, we introduce \model, a Video Editor with Grounded Generation from Instructions. 
\model unifies video concept grounding and editing without relying on additional layout, mask guidance, or intermediate caption~\cite{yu2024zero,yoon2024raccoon,ku2024anyv2v,lian2023llm,yu2023inpaint, li2025training}. Instead, we formulate the problem as end-to-end grounded generation in pixel space, using only a diffusion loss.
Specifically, given a video and a text query, \model first leverages an MLLM to interpret complex instructions, generating frame-wise conditions. 
Unlike prior methods~\cite{yoon2024raccoon,lin2023videodirectorgpt} that use discrete text tokens as conditions, which disconnect the pipeline and block gradient propagation, \model employs continuous, learnable task query embeddings per frame.
This enables end-to-end training and effectively captures grounded task representations for diffusion model conditioning.
To handle diverse tasks and accurately interpret complex queries, we employ a curriculum learning strategy that begins by aligning MLLMs with diffusion models using massive paired instructional image editing data and then fine-tuning the model end-to-end on high-quality multitask video data to adapt video. 
Unlike tool-use methods~\cite{yoon2024raccoon,fei2024vitron,ku2024anyv2v}, \model formulates both video grounding and instructional editing in the same video-to-video task formulation, enabling efficient handling through a unified single model. 
To further support end-to-end training, we introduce a novel automatic instructional video data generation pipeline that lifts high-quality instructional image editing data into the video domain using image-to-video and video evaluation tools.

\begin{figure*}
    \centering
    \includegraphics[width=0.83\linewidth]{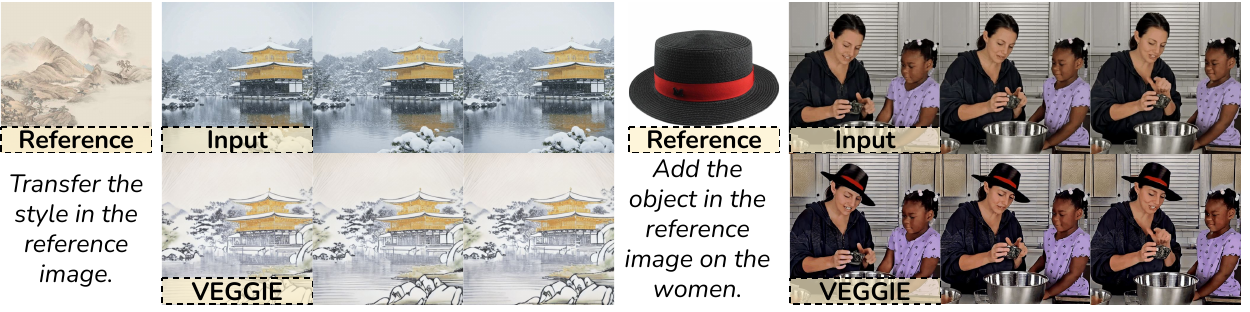}
    \caption{Multimodal instruction following emerges in \model, allowing for style transfer or object addition from reference images.}
    \label{fig:main_multimodal}
\end{figure*}

\begin{figure*}
    \centering
    \includegraphics[width=0.83\linewidth]{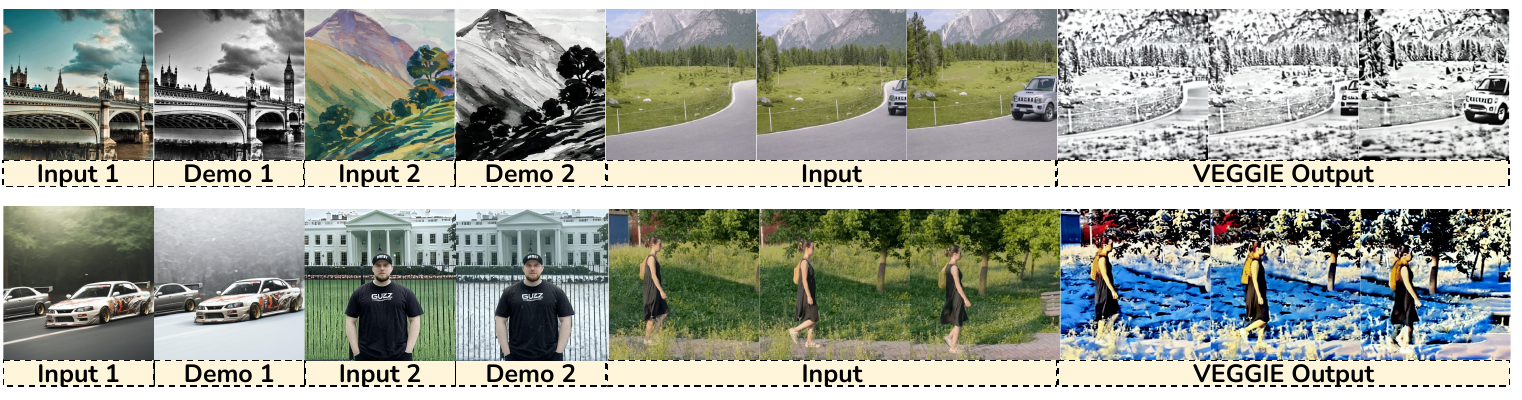}
    \caption{In-context editing emerges in \model, allowing for few-shot learning of editing tasks with paired image demonstrations.}
    \label{fig:main_icl}
\end{figure*}

Existing video editing benchmarks do not provide wide and uniform coverage of diverse editing skills~\cite{wu2023cvpr,sun2024bench}. 
To address this gap, we contribute \benchmark, an instructional video editing benchmark that spans 8 editing skills: concept addition, removal, object changing, environment background changing, visual feature changing, stylization, object grounding, and reasoning segmentation. 
Each skill is evaluated using a dedicated suite of metrics.
We assess the proposed \model alongside 6 baselines on \benchmark. 
\model demonstrates strong performance across diverse editing skills, outperforming the best instructional baseline as a versatile, all-in-one model, while other models struggle with multi-tasking. 
Additionally, \model excels in video object grounding and reasoning segmentation tasks, where other baselines fall short.
We show further analysis of how multi-task learning enhances our framework and highlight applications such as \textbf{zero-shot multimodal instructional following} (\cref{fig:main_multimodal}) and \textbf{few-shot in-context editing} (\cref{fig:main_icl}).
Our contributions are summarized as follows:
\begin{itemize}[leftmargin=10pt, topsep=1pt, noitemsep]
    \item We propose \model, an end-to-end model that integrates an MLLM and a VidDM. \model is a versatile framework that handles diverse instructional requests for editing and grounding various video concepts. 
    Unlike existing work that achieves multitasking via tool use, \model unifies diverse tasks in a single model, thus simplifying the training with only diffusion loss. 
    \item We propose a data synthesis pipeline, scaling high-quality instructional video editing data for future work.
    \item We propose \benchmark, an instructional video editing benchmark that spans 8 editing skills with dedicated metrics for each skill.
    \item \model achieves strong performance across diverse editing skills compared with SoTA methods, and shows potentials for multimodal instruction and in-context following. 
\end{itemize}

%% file: sec/2_related_work.tex
\section{Related Work}
\label{sec:background}

\noindent\textbf{Instructional Video Editing}
Video Diffusion Models (VidDMs)~\citep{yan2021videogpt, hong2023cogvideo, esser2023structure, mei2023vidm, chai2023stablevideo, ceylan2023pix2video, blattmann2023svd, wu2023tune, brooks2024video}, enable high-quality video generation across a wide range of video concepts. 
Building on these advances, video editing methods have emerged as tools for video design, allowing users to manipulate video concepts such as adding, removing, altering objects and style translation~\citep{geyer2023tokenflow,qi2023fatezero,wang2024videocomposer,yu2023inpaint,zhang2023avid}.
To enhance user experiences, instructional video editing methods~\cite{qin2024instructvid2vid,zhang2024effived, xing2023vidiff} have been developed, using triples of video instructions, source videos, and target videos for training. 
These methods demonstrate limited performance when complex multimodal reasoning is required, as noted by previous research on instructional image editing~\cite{fu2023guiding}.
Moreover, they struggle with diverse editing tasks, from addition and deletion to stylization. 
For example, LGVI~\cite{wu2024towards} is primarily designed for removal tasks, while TokenFlow~\cite{geyer2023tokenflow} struggles with local edits such as adding, removing, or changing objects.
We address this limitation with pixel-level multitasking fine-tuning on well-curated instructional video editing datasets covering various grounding and editing skills.

\noindent\textbf{Video Grounding and Segmentation}
Visual grounding requires models to connect language to its corresponding visual concept in the visual context~\cite{li2022grounded,ma2023world}.
This is commonly evaluated via the language-guided semantic localization tasks, ranging from simple referring expressions in RefCOCO series~\cite{yu2016modeling,mao2016generation} and their generalized variant~\cite{liu2023gres} that takes no-target and multi-target into account.
Recently, grounded multimodal large language models (MLLMs) are trained for object grounding to bounding boxes~\citep{pi2023detgpt,chen2023shikra,zhao2023bubogpt,you2023ferret,zhang2024ferret,peng2024grounding} and segmentation masks~\citep{zhang2023llava,xia2023gsva,rasheed2023glamm,zhang2024groundhog} using text-image pairs with fine-grained annotations linking phrases to entities.
These models unlock the potential of reasoning segmentation~\citep{lai2023lisa,deng2024motion}, bringing language-informed reasoning into semantic segmentation.
Instead of using dedicated object detection or segmentation modules, we achieve video grounding through end-to-end training with only diffusion loss.

%% file: sec/3_method.tex
\section{Our Method: \textbf{\model}}

\begin{figure*}
    \centering
    \includegraphics[width=0.95\linewidth]{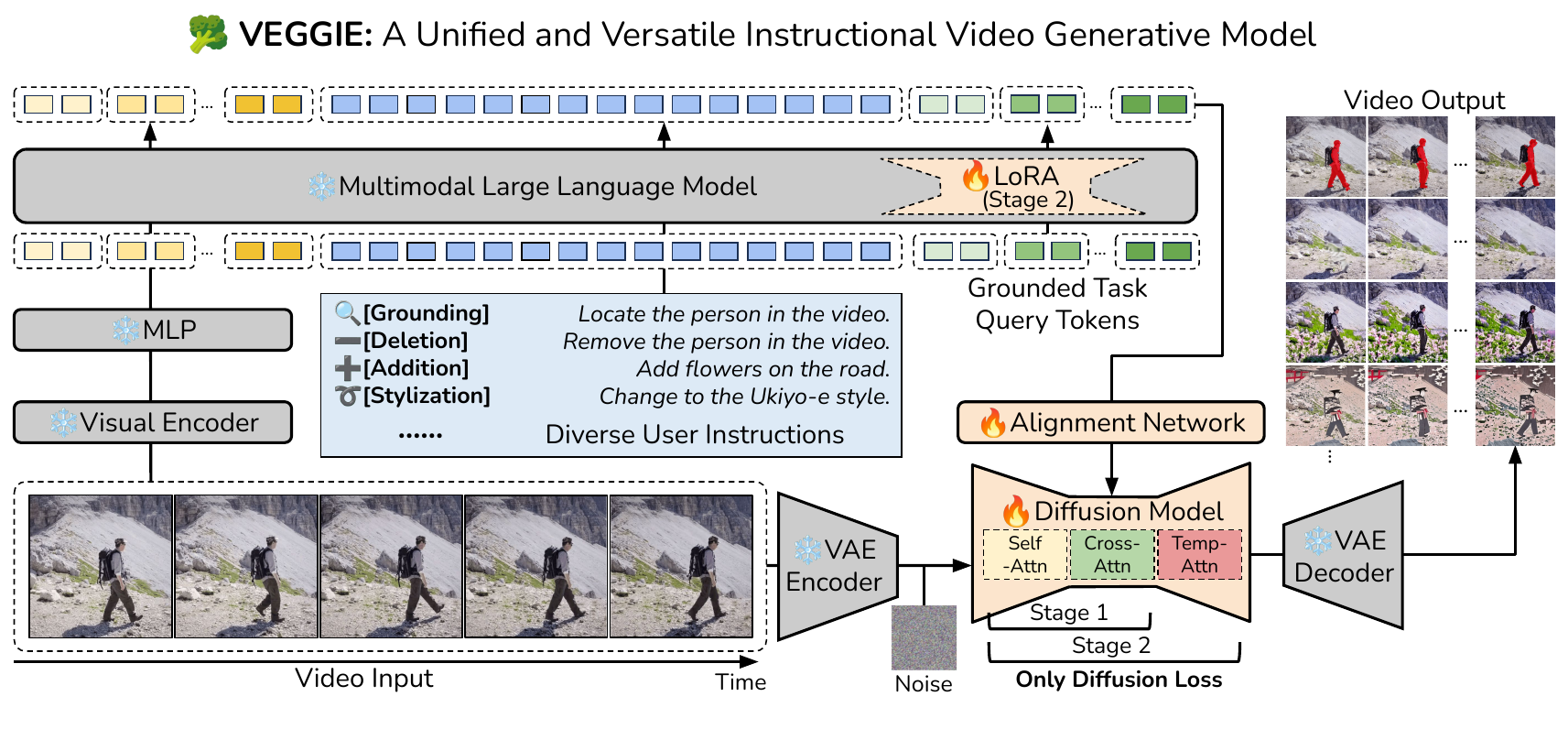}
    \vspace{-5pt}
    \caption{Overview of our proposed end-to-end \model framework. 
    Our Multimodal Large Language Model first understands input video frames and diverse user instructions, then it generates frame-wise reasoning queries that maintain per-frame editing conditions for the video diffusion model. The video diffusion model will render the MLLM-generated conditions to the pixel space for diverse tasks, including video editing, video grounding, and video reasoning segmentation with questions. We only apply diffusion loss for the whole pipeline training. 
    }
    \label{fig:main}
\end{figure*}

In this paper, we introduce a Video Editor with Grounded Generation from Instructions (\textbf{\model}), a \textbf{unified and versatile generative video model}.
It combines the complex instruction understanding and reasoning capabilities of MLLMs with the generative capacity of VidDMs. The model is trained end-to-end \textbf{with diffusion loss only}.
\model efficiently handles diverse user inputs, including direct instructions, complex questions requiring in-depth reasoning, and multimodal conditioning. 
It performs various pixel-level manipulations, enabling tasks such as video concept addition, removal, changing, stylization, grounding, and reasoning segmentation based on user instructions.
We elaborate on the model design (\cref{sec:model}), training and inference process (\cref{sec:training}), and data curation (\cref{sec:data}),.

\subsection{Model Architecture}
\label{sec:model}

\model consists of four main components (see~\cref{fig:main}): 
(1) a multimodal large language model, 
(2) a set of learnable grounded task queries, 
(3) an alignment network (single-layer MLP) that projects the MLLM output into the condition space of the diffusion model, and 
(4) a video diffusion model initialized from an instructional image editing model~\cite{zhang2024magicbrush}. 
Our model first generates latent conditions for target video frames by querying multimodal context using an MLLM, then renders these conditions at the pixel level through a video diffusion model, as detailed below.

\noindent\textbf{MLLM for Generating Grounded Task Guidance.} 
As illustrated in the left of~\cref{fig:main}, given a video consisting of a sequence of frames $V=[f_1,\dots,f_n]$, where $n$ is the frame number of the given video, a user instruction/question $I$, our goal is to obtain the response $\widehat{V}=[\widehat{f_1},\dots,\widehat{f_n}]$ at pixel space that faithfully reflects user instruction about the given video. 
The MLLM module processes both the input video $V$ and a user instruction $I$ to generate a sequence of grounded task tokens per frame: $C = [c_1,\dots,c_n]$, which are input and output in parallel.
These tokens serve as task guidance and implicitly encode the target manipulation, such as object attributes, spatial relationships, or style transfer parameters. 
The MLLM ensures the model captures both explicit user instructions and implicit reasoning needs.

\noindent\textbf{VidDM for Rendering MLLM Guidance at Pixel Space.}
As illustrated in the right of~\Cref{fig:main}, the VidDM takes the original video $V$ and the grounded task tokens $C$ as conditions to synthesize the target video $\widehat{V}$. The original video is concatenated with the noise volume, and the task tokens are input to the cross-attention.
With grounded task guidance in denoising steps, the generation process ensures that the output faithfully follows user instructions while preserving the video's structure and motion dynamics. 
Through iterative denoising, it refines each frame while maintaining temporal consistency, applying pixel modifications coherently for a smooth and visually consistent output video $\widehat{V}$.

\subsection{Curriculum Learning from Images to Videos}
\label{sec:training}

Training the model directly on video tasks presents two key challenges: 
(1) misalignment between MLLM and diffusion model representations, making it difficult for the diffusion model to interpret MLLM-generated task queries with limited fine-tuning data, and 
(2) the diffusion model's lack of multitasking capability, even for image tasks, due to insufficient training on diverse tasks. 
Our initial experiments also found the model collapsed when the whole pipeline was directly trained with all data.
These challenges/observations underscore the need for pre-alignment between MLLM and the diffusion model to enable seamless adaptation from language-space task queries to pixel-space modifications.
To this end, we adopt a two-stage curriculum learning strategy for the proposed \model framework. 

\noindent\textbf{Stage 1: Aligning Diffusion and Language Spaces.}
In the first stage, we align the diffusion model with the MLLM using large-scale image-level instructional editing data. The MLLM remains frozen while we update the alignment network, grounded task queries, and diffusion UNet. 
This process fine-tunes the diffusion model weights to align with the language space, enabling the model to interpret MLLM-generated guidance and translate user instructions into pixel-level edits while preserving the MLLM’s strong ability to understand instructions and user intentions.

\noindent\textbf{Stage 2: Enhancing Temporal Consistency and Dynamics.}
With the MLLM and diffusion model aligned, fine-tuning diverse instructional video editing data becomes more effective for improved instruction following at pixel-space including temporal consistency, dynamic coherence, and editing faithfulness. 
In this stage, we fine-tune the framework with the MLLM, including the alignment network, grounded task queries, and all 3 dimensions in diffusion UNet, \textbf{end-to-end} with carefully curated multitasking instructional video editing data. 
Following prior work ~\cite{guo2023animatediff, wu2023tune}, we inflated the 2D UNet from Stage 1 with temporal attention layers for video adaptation. 
For both stages 1 and 2, we optimize the framework with a \textbf{single diffusion loss}, enabling unified learning for improved instructional video editing performance while maintaining simplicity and efficiency.

\noindent\textbf{Classifier-Free Guidance during Testing.} 
We employ classifier-free guidance to balance quality and diversity in diffusion-generated samples. Following prior work ~\cite{brooks2023instructpix2pix, fu2023guiding}, we apply classifier-free guidance to instructional visual editing considering two conditions: the grounded task tokens and the original video. 
To obtain unconditional guidance, we set null values ($\varnothing$) for both task tokens and input video. In this case, our score estimate is:
\vspace{-5pt}
\begin{equation*}
\begin{split}
    \tilde{e_{\theta}}(z_t, c_T, c_V) = &\: e_{\theta}(z_t, \varnothing, \varnothing) \\ 
    &+ g_T \cdot (e_{\theta}(z_t, c_V, c_T) - e_{\theta}(z_t, c_V, \varnothing))
    \\ &+ g_V \cdot (e_{\theta}(z_t, c_V, \varnothing) - e_{\theta}(z_t, \varnothing, \varnothing)),
    \label{eq:cfg}
\end{split}
\end{equation*}

\vspace{-5pt}
\noindent
where $\theta$ represents the model parameters, $C_T$ and $C_V$ denote the task tokens and video conditions, $\varnothing$ is the null value, $z_t$ is the noised latent at timestamp $t$, and $g_T$ and $g_V$ are the task guidance and video guidance scales, respectively. 
More training details are included later in Appendix.

\input{materials/tab_data_short}

\subsection{Data Curation Pipeline} 
\label{sec:data}

Existing video editing models, both instructional and non-instructional, struggle with diverse editing skills due to the lack of high-quality multitasking fine-tuning data. 
In this section, we introduce our data curation strategy to support \model in achieving versatile video editing skills. 
As listed in~\cref{tab:data_summary}, we collect 3.4M image and 133.9K video data from diverse sources to support our \model curriculum learning as discussed in~\cref{sec:training}.
We create our training dataset from two sources: (1) collecting existing image and video data and converting it into an instructional editing format, and (2) synthesizing new instructional video editing samples using existing datasets and generative models.

\noindent\textbf{Collecting Diverse Multitask Image and Video Data.}
We bring together instructional editing data from both image (Seed-Data-Edit~\cite{ge2024seed}, MagicBrush~\cite{zhang2024magicbrush}, EraseDraw~\cite{canberk2024erasedraw}) and video (InstructV2V~\cite{cheng2023consistent}, VPLM~\cite{yoon2024raccoon}) sources. 
These datasets provide pairs of original and edited visual contents with user instructions.
The tasks include adding, removing, and changing objects, stylizing, and performing global/local edits. 
Beyond editing datasets, we incorporate segmentation data at both the image level (gRefCoCo~\cite{liu2023gres} and PhraseCut~\cite{wu2020phrasecut}) and the video level (RVoS and MeViS). These segmentation tasks are reformulated as color-filling challenges, which guide the model in learning referring grounding (i.e., understanding which object or region to edit) and strengthen its conceptual learning. 
To further unlock complex instruction understanding via MLLM, we include data that requires more advanced reasoning and implicit referencing. 
Specifically, we include: reasoning segmentation (LISA~\cite{lai2024lisa}), reasoning editing (SmartEdit~\cite{huang2024smartedit}), interactive video inpainting (LGVI~\cite{wu2024towards}), and motion-grounded video reasoning (GroundMoRe~\cite{deng2024groundmore}). 
These tasks help \model learn implicit references and reasoning.

\begin{figure}
    \centering
    \includegraphics[width=0.85\linewidth]{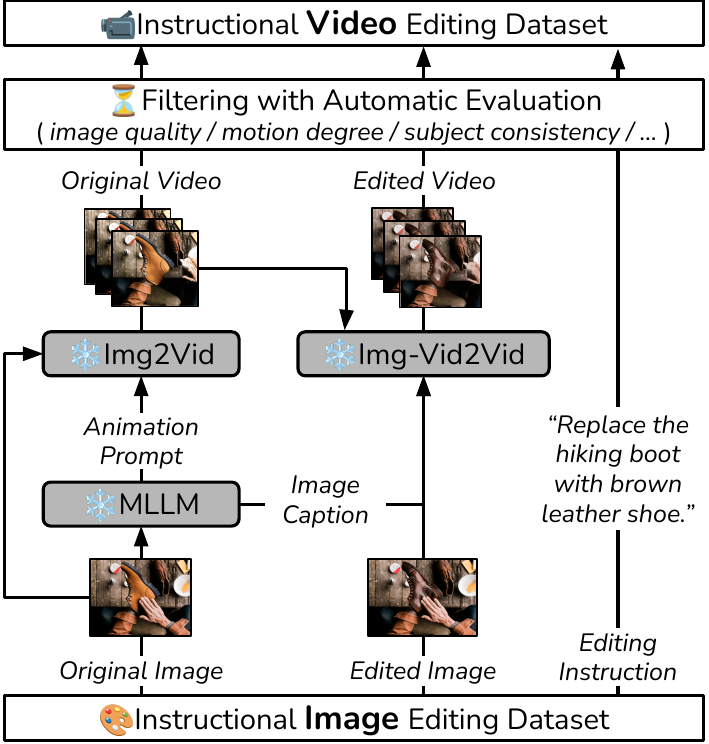}
    \vspace{-5pt}
    \caption{Our data generation pipeline for synthetic instructional video editing data. It injects dynamics into well-constructed instructional image editing datasets via the Image-to-Video (I2V) Model, and generates paired video data for instruction editing. \vspace{-12pt}}
    \label{fig:data_pipeline}
\end{figure}

\noindent\textbf{Synthesizing Instructional Video Editing Data via Image-to-Video Animation.}
Recent methods~\cite{qin2024instructvid2vid, brooks2023instructpix2pix} generate synthetic instructional video-editing data by first creating text instructions with LLM, then getting edited videos via T2V models and prompt-to-prompt editing~\cite{hertz2022prompt}. 
While these methods adapt image-based editing pipelines~\cite{cheng2023consistent} to videos, the generated data suffer from temporal-consistency issues.
To address this gap, we propose a novel \textbf{image-to-video animation strategy} that leverages the abundance of high-quality \emph{image}-level instructional editing datasets~\cite{zhang2024magicbrush, wei2024omniedit}, which provide well-annotated instructions, paired edited images, and well-organized editing skill categories.
As illustrated in~\cref{fig:data_pipeline}, given an original image $I$, an edited image $\bar{I}$, and an instruction from an instructional image editing dataset~\cite{wei2024omniedit}, our approach involves three key steps. 
First, we use an offline MLLM~\cite{qwen2,Qwen2VL} to generate an image caption and an animation prompt that describes plausible motion within the image. 
Next, an image-to-video (I2V) model animates the image into a video $V$. 
Finally, we generate the corresponding edited video $\bar{V}$ using a first-frame-conditioned video editing model~\cite{ku2024anyv2v}, leveraging $\bar{I}$ as a strong prior to ensure consistent edits across frames.
Finally, to ensure data quality, we evaluate each original-edited video pair with automatic video quality evaluation metrics~\cite{huang2024vbench}, which assess the generated videos from diverse dimensions, e.g., motion smoothness, image quality, and background consistency. 
This pipeline transforms carefully curated image-based datasets into instructional video-editing resources while preserving the precision of the original edits. 
As a result, our data method expands the availability of high-quality synthetic video-editing data, supporting a wider range of editing tasks in our end-to-end unified framework. 
More details on data generation, prompting, examples, and pre/post-processing are in the Appendix. 

%% file: materials/tab_data_short.tex
\begin{table}
    \centering
    \scalebox{0.9}{
    \begingroup
    \setlength{\tabcolsep}{3pt}
    \renewcommand{\arraystretch}{0.8}
    \hspace{-10pt}
    \begin{tabular}{llccccr}
    \toprule
    \multicolumn{1}{l}{\textbf{Type}} & 
    \multicolumn{1}{l}{\textbf{Source}} & 
    \multicolumn{1}{l}{\textbf{R.}} &
    \multicolumn{1}{l}{\textbf{E.}} &
    \multicolumn{1}{l}{\textbf{G.}} &
    
    \multicolumn{1}{c}{\textbf{\# Img./Vid.}} & 
    \multicolumn{1}{r}{\textbf{\# Ins.}} \\ \midrule

    \multirow{9}{*}{Video} 
    & ROVI~\cite{wu2024towards} & \cmark & \cmark & \cmark & 4.3K & 27.4K \\
    & VPLM~\cite{yoon2024raccoon} & & \cmark  & & 4.3K & 5.5K \\
    & GroundMoRe~\cite{deng2024groundmore} & \cmark &  & \cmark & 1.3K & 5.5K \\
    & RVoS~\cite{ventura2019rvos} &  &  & \cmark & 1.9K & 6.1K \\
    & MeViS~\cite{ding2023mevis} & & & \cmark & 1.5K & 17.1K \\ 
    & InstructV2V~\cite{cheng2023consistent} & & \cmark & & 68.3K & 68.3K \\ 

    & \textbf{\dataset} (Ours) & & \cmark & & 4.0K & 6.2K \\ \midrule

    \rowcolor[HTML]{D8D8D8}
    \multicolumn{5}{l}{Total} &  & 136.1K \\
    
    \midrule
    
    \multirow{6}{*}{Image} 
    & Seed-Data-Edit~\cite{ge2024seed} & & \cmark & & 3M & 3M \\
    & LISA~\cite{lai2024lisa} &  \cmark &  & \cmark & 0.2K & 1.3K \\
    & gRefCoCo~\cite{liu2023gres} & & & \cmark & 13.6K & 73.4K \\
    & PhraseCut~\cite{wu2020phrasecut} & & & \cmark & 310.8K & 310.8K \\
    & EraseDraw~\cite{canberk2024erasedraw} & & \cmark & & 64.9K & 42.4K \\ 
    & MagicBrush~\cite{zhang2024magicbrush} & & \cmark & & 9.3K & 9.3K \\
    & SmartEdit~\cite{huang2024smartedit} & & \cmark & & 0.5K & 0.9K \\
    
    \midrule
    \rowcolor[HTML]{D8D8D8}
    \multicolumn{5}{l}{Total} &  & 3438.1K \\
    \bottomrule
    \end{tabular}
    \endgroup}
    \vspace{-5pt}
    \caption{Summary of our data for training. \textbf{R.}: Reasoning, \textbf{E.}: Editing, \textbf{G.}: Grounding. \textbf{\# Img/Vid}: the number of images/videos, and \textbf{\# Ins.}: the number of instruction-image/video pairs.\vspace{-10pt}}
    \label{tab:data_summary}
\end{table}

%% file: sec/4_exp.tex
\section{Experiments}
\begin{figure*}
    \centering
    \includegraphics[width=0.95\linewidth]{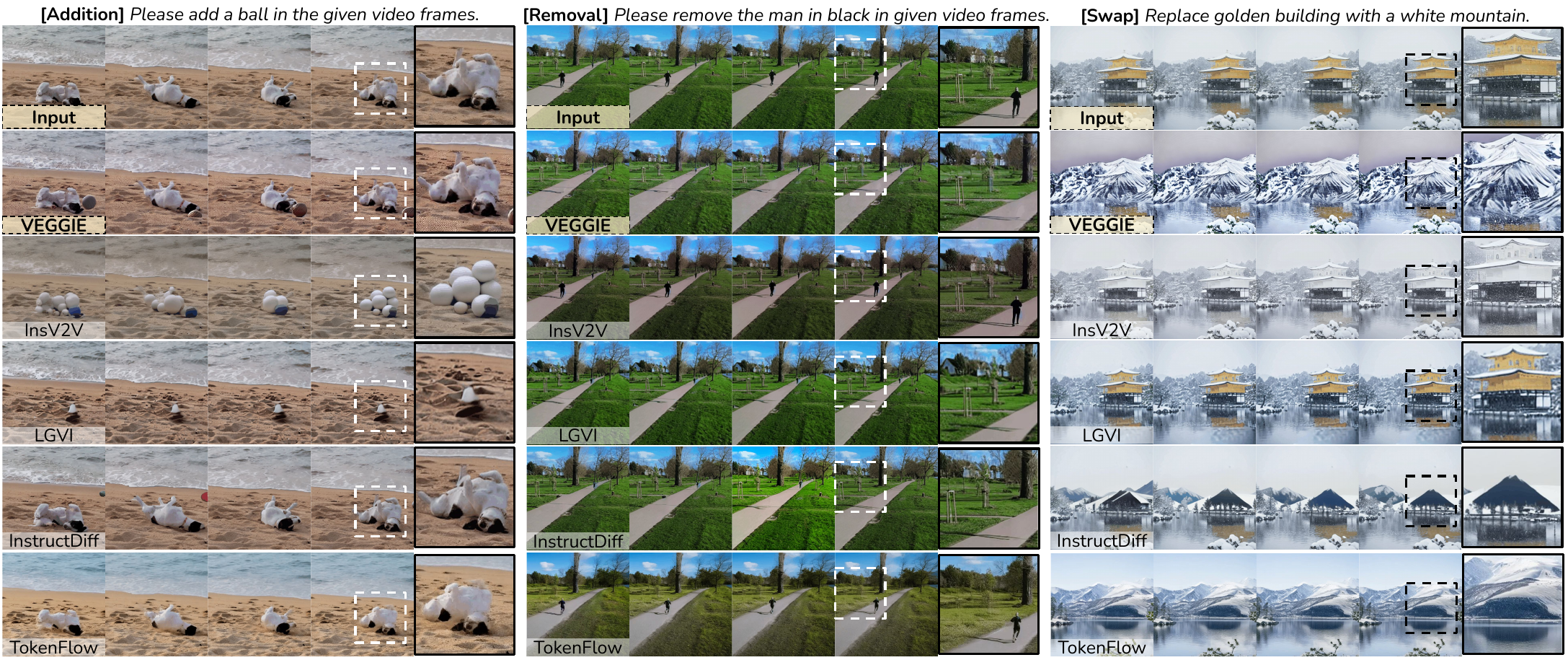}
    \includegraphics[width=0.95\linewidth]
    {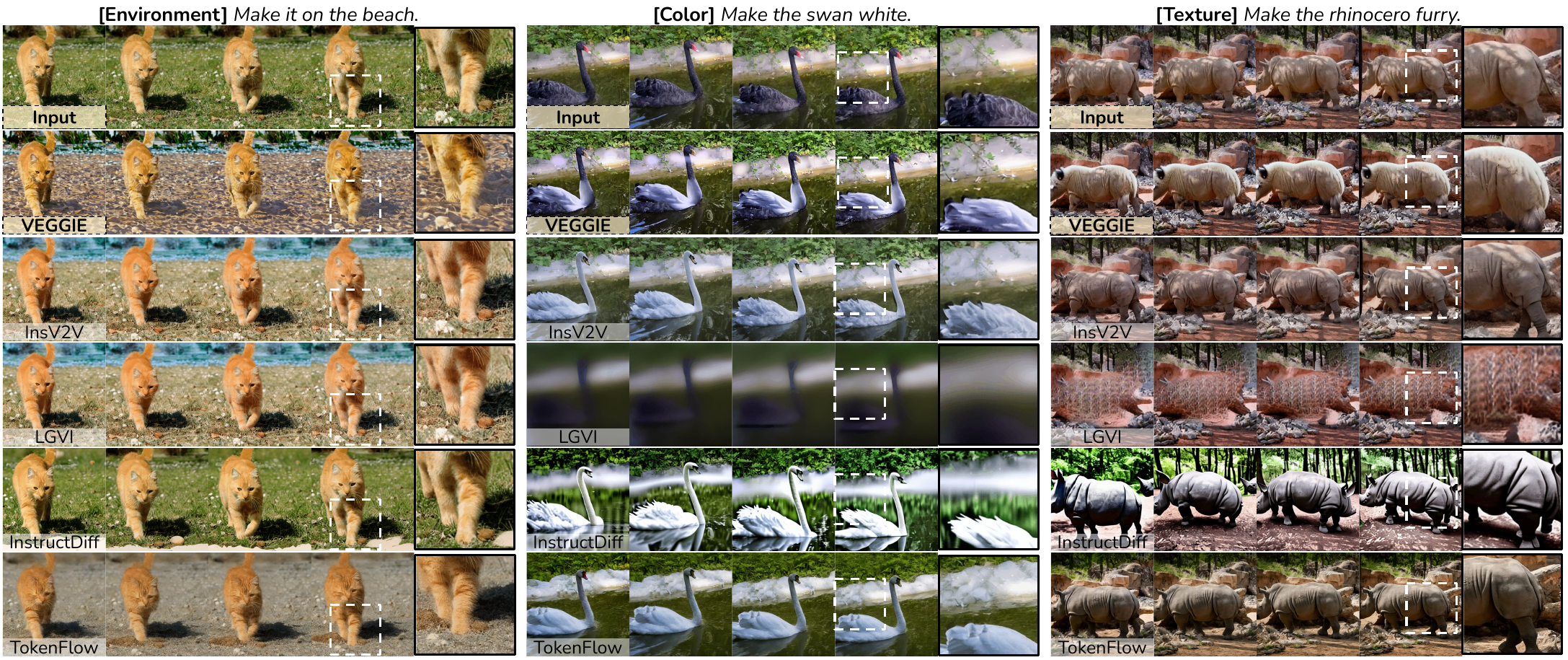}
    \includegraphics[width=0.95\linewidth]
    {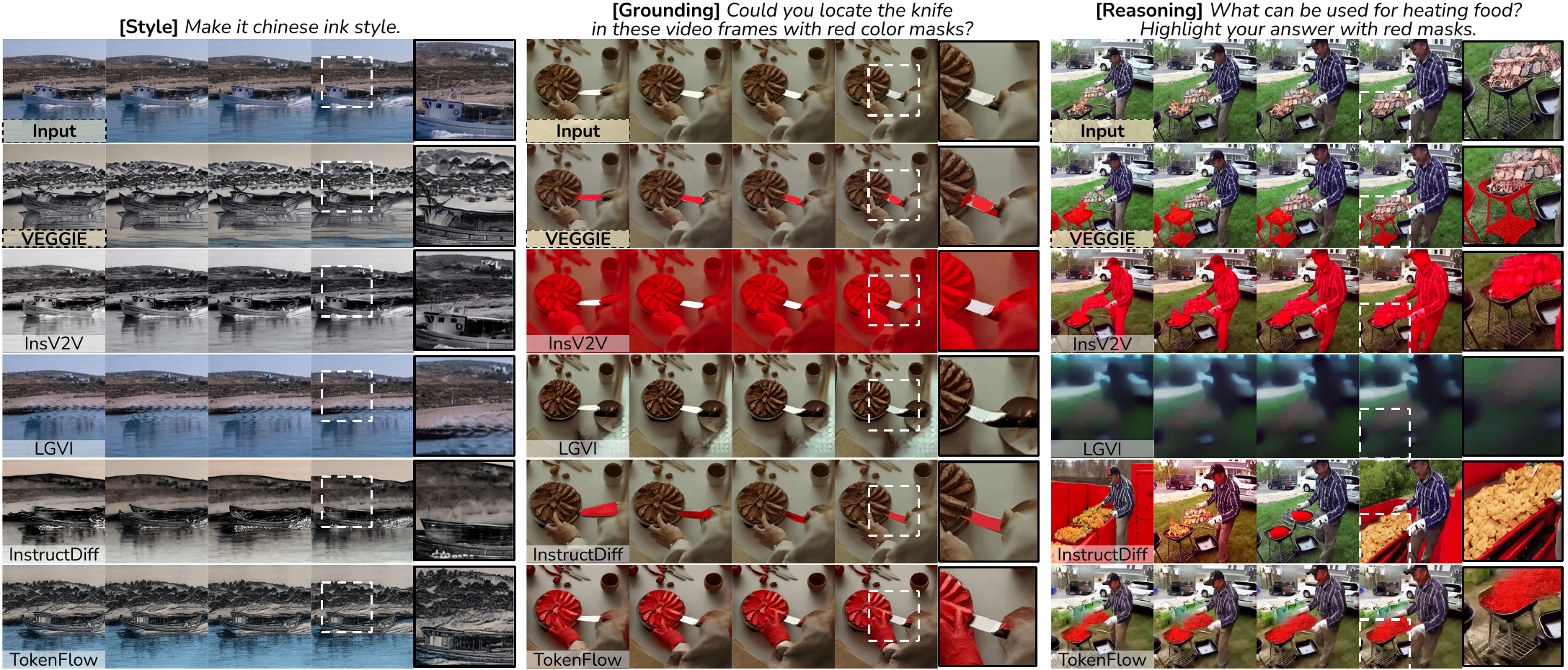}
    \caption{Qualitative comparison of editing results across 8 different abilities (splitting visual features into color and texture). We provide zoom-in details for a more detailed comparison. Best viewed in color. More in Appendix.}
    \label{fig:main_results}
\end{figure*}

\input{materials/vebench}

We first introduce the \benchmark Benchmark and then demonstrate the superiority of \model across diverse video instructional editing skills.
More experiments, visualization, and implementation details are in the Appendix.


\subsection{\textbf{\benchmark} and Metrics}
\label{sec:benchmark}

As no existing benchmark is designed for fine-grained instructional video editing skills, we manually collect and annotate \benchmark, containing 132 video-instruction pairs that balanced cover 8 different video generative skills (15$-$20 for each).
Beyond standard metrics, including text-to-video alignment (CLIP-Text~\cite{radford2021learning}), video smoothness (CLIP-F~\cite{radford2021learning}), and image quality (MUSIQ~\cite{ke2021musiq}), we also first introduce \textbf{MLLM-as-a-Judge} to give a holistic evaluation score according to the given original video, edited video, and user instruction. 
It is achieved by prompting GPT-4o~\cite{gpt4o} to evaluate whether the requested semantic change has been fulfilled, using a scale from 1 to 10.
For addition and removal, we also introduce an object detector (GroundingDiNo~\cite{liu2023grounding}) to detect if the object is added/removed faithfully. 
For grounding and reasoning segmentation, we following video grounding tasks~\cite{deng2024groundmore,refdavis,refytvos,ding2023mevis} and adopt the Jaccard index ($\mathcal{J}$)~\cite{jaccard1912distribution}, F-measure ($\mathcal{F}$)~\cite{dice1945measures}, and their mean ($\mathcal{J\&F}$).
We also compute SSIM between the generated video and the original video masked with GT masks. More evaluation/metrics details are included in Appendix.


\begin{table}[]
    \scalebox{0.9}{
    \begingroup
    \setlength{\tabcolsep}{5pt}
    \renewcommand{\arraystretch}{0.8}
    \begin{tabular}{lcc}
    \toprule
    \textbf{Settings} & \multicolumn{1}{c}{\textbf{Removal} (FVD $\downarrow$)} & \multicolumn{1}{c}{\textbf{Grounding} (SSIM $\uparrow$)} \\ \midrule
    Grd.-only & - & 52.34 \\
    Rmv.-only & 1098.52 & - \\ \midrule
    Mixed & \textbf{987.80} & \textbf{55.21}\\ \bottomrule
    \end{tabular}
    \endgroup}
    \caption{An ablation study on whether multi-task learning provides transferable benefits that enhance performance across tasks. We focus on removal and grounding tasks as representative examples.}
    \label{tab:ablate}
\end{table}

\begin{figure}
    \centering
    \includegraphics[width=0.9\linewidth]{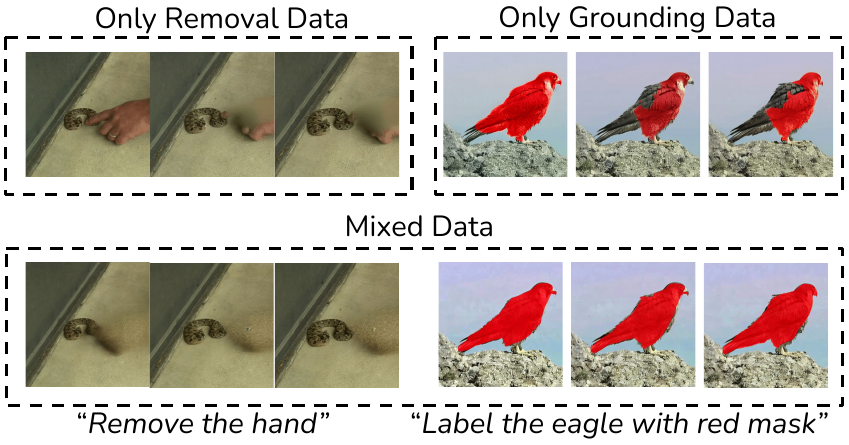}
    \caption{Comparison between single- and multi-skill models with different data training. We find tasks can help each other. \vspace{-5pt}}
    \label{fig:ablate}
\end{figure}

\subsection{Experimental Results}

\noindent\textbf{Instructional Video Editing over Diverse Skills.}
As shown in~\cref{tab:main}, we evaluate 7 different models on \benchmark across 8 distinct editing skills. Overall, \model demonstrates the best performance among instructional video editing models.
Compared to \model, non-instructional models often struggle with concept removal and addition. This limitation arises because these models rely on attention control or additional conditions (e.g., depth maps) that impose strong priors, constraining the model and making object addition or removal challenging.
We also observe that InsV2V achieves high scores in quality and smoothness metrics, but underperforms in alignment and MLLM judgment, which demand faithful semantic changes. Qualitative examples in Fig.~\ref{fig:main_results} illustrate that InsV2V often makes minimal changes to the input video, resulting in high video quality but unfaithful outputs.
In contrast, \model strikes a better balance, delivering both high-quality visuals and accurate semantic alignment with the intended edits.


\noindent\textbf{Can Multi-Task Help Each Other?}
To test the previous hypothesis, we train our model on the VPLM~\cite{yoon2024raccoon} dataset, which includes paired grounding and removal tasks (approximately 5.5K samples for each task). 
We focus on these tasks as representative examples due to their straightforward evaluation against ground truth.
As shown in~\Cref{tab:ablate}, multitask training yields a lower FVD score and a higher SSIM score, demonstrating that learning to locate and remove a video concept can mutually reinforce performance.
We show an example in Fig.~\ref{fig:ablate}.
However, this conclusion only holds with a balanced data combination. 
We also observe that an excessive amount of grounding data can introduce more artifacts and negatively impact visual editing skills.

\noindent\textbf{Emergent Zero-shot Multimodal Instruction Following.} 
We also highlight the emergent behavior of \model on multi-modal instruction following, even without dedicated training data for this specific editing instruction. 
Notably, \model demonstrates the ability to perform zero-shot multimodal instructional video editing. As illustrated in Fig.~\ref{fig:main_multimodal}, \model can transfer styles or add objects from a reference image into the input video based on instructions.

\noindent\textbf{Emergent Few-shot In-Context Editing.}
As shown in Fig.~\ref{fig:main_icl}, \model can effectively utilize a few example image pairs to transfer the intended editing changes seamlessly to the input video.
We observe that \model exhibits in-context learning for image editing without the need for language instructions. Instead, it uses image pairs as examples to infer and apply the desired editing intention directly.

%% file: materials/vebench.tex
\definecolor{ggg}{gray}{0.5}

\begin{table*}[t]
\centering
\small
\scalebox{0.82}{
\begingroup
\setlength{\tabcolsep}{4.5pt}
\renewcommand{\arraystretch}{0.75}
\hspace{-10pt}
\begin{tabular}{lcccccccc}
\toprule
\multirow{2}{*}{\textbf{Methods}} 
& \multicolumn{3}{c}{\textbf{Non-Instructional Editing Model}} 
&  
& \multicolumn{4}{c}{\textbf{Instructional Editing Model}} 
\\ \cmidrule(lr){2-4} \cmidrule(lr){6-9} 
& \textbf{VidToMe}~\cite{li2024vidtome}
& \textbf{TokenFlow}~\cite{geyer2023tokenflow} 
& \textbf{Flatten}~\cite{cong2024flatten} 
&  
& \textbf{InstructDiff}~\cite{geng2024instructdiffusion}    
& \textbf{LGVI}~\cite{wu2024towards}    
& \textbf{InsV2V}~\cite{cheng2023consistent}   
& \textbf{\model} (Ours)    
\\ 
\textbf{Generator} & (SD1.5) & (SD2.1) & (SD2.1) & & (SD1.5) & (SD1.5) & (SD1.5) & (SD1.5) \\
\midrule
\rowcolor[HTML]{fadbd8} 
\multicolumn{9}{c}{\textit{Concept Addition}} \\ 
MLLM-Judge ($\uparrow$) 
& \textcolor{ggg}{5.00}
& \textcolor{ggg}{5.80}
& \textcolor{ggg}{6.62}
&  
& \underline{7.26} 
& 2.73  
& 5.69 
& \textbf{7.44} 
\\ 
Alignment ($\uparrow$) 
& \textcolor{ggg}{27.80} 
& \textcolor{ggg}{29.30} 
& \textcolor{ggg}{28.22} 
&  
& 28.10 
& 25.06 
& \underline{28.27} 
& \textbf{29.27} 
\\ 
Smoothness ($\uparrow$) 
& \textcolor{ggg}{96.79}
& \textcolor{ggg}{97.26}
& \textcolor{ggg}{95.74}
&  
& 93.66 
& \underline{96.09} 
& \textbf{96.94} 
& 94.93 
\\ 
Quality ($\uparrow$) 
& \textcolor{ggg}{62.81}
& \textcolor{ggg}{65.62}
& \textcolor{ggg}{55.45}
&  
& 54.08 
& 41.59 
& \underline{56.24} 
& \textbf{61.31} 
\\ 
Detection ($\uparrow$) 
& \textcolor{ggg}{47.98}
& \textcolor{ggg}{49.53}
& \textcolor{ggg}{49.74}
&  
& \underline{55.36}
& 14.42
&48.01
& \textbf{57.96}
\\ 
\rowcolor[HTML]{e8daef} 
\multicolumn{9}{c}{\textit{Concept Removal}} \\ 
MLLM-Judge ($\uparrow$) 
& \textcolor{ggg}{2.60}
& \textcolor{ggg}{3.73}
& \textcolor{ggg}{4.46}
&  
& \underline{6.12} 
& \textbf{6.59}  
& 2.78 
& 5.07 
\\ 
Alignment ($\uparrow$) 
& \textcolor{ggg}{75.01}
& \textcolor{ggg}{75.99}
& \textcolor{ggg}{78.40}
&  
& 75.51
& \textbf{75.67}
& 74.41
& \underline{75.63}
\\ 
Smoothness ($\uparrow$) 
& \textcolor{ggg}{96.13}
& \textcolor{ggg}{96.47}
& \textcolor{ggg}{95.82}
&  
& 91.83 
& \textbf{97.03}
& \underline{96.99}
& 95.04 
\\ 
Quality ($\uparrow$) 
& \textcolor{ggg}{66.32}
& \textcolor{ggg}{71.52}
& \textcolor{ggg}{50.77}
&  
& \underline{55.08}
& 42.31
& \textbf{58.79}
& 50.99 
\\ 
Detection ($\uparrow$) 
& \textcolor{ggg}{34.31}
& \textcolor{ggg}{55.16}
& \textcolor{ggg}{70.91}
&  
& 64.81
& \textbf{78.40}
& 25.64
& \underline{70.22}
\\ 
\rowcolor[HTML]{d6eaf8} 
\multicolumn{9}{c}{\textit{Object Changing}} \\ 
MLLM-Judge ($\uparrow$) 
& \textcolor{ggg}{5.00}
& \textcolor{ggg}{6.53}
& \textcolor{ggg}{7.37}
&  
& \textbf{7.00} 
& 2.06  
& 6.60 
& \underline{6.63} 
\\ 
Alignment ($\uparrow$) 
& \textcolor{ggg}{25.69}
& \textcolor{ggg}{28.76}
& \textcolor{ggg}{27.06}
&  
& \underline{27.36}
& 22.17
& 26.60
& \textbf{27.77} 
\\ 
Smoothness ($\uparrow$)  
& \textcolor{ggg}{96.23}
& \textcolor{ggg}{97.21}
& \textcolor{ggg}{96.13}
&  
& 92.07
& \underline{95.66}
& \textbf{96.74}
& 95.44
\\ 
Quality ($\uparrow$) 
& \textcolor{ggg}{64.06}
& \textcolor{ggg}{69.97}
& \textcolor{ggg}{59.37}
&  
& 55.01
& 38.20
& \textbf{60.90}
& \underline{58.15}
\\ 
\rowcolor[HTML]{d0ece7} 
\multicolumn{9}{c}{\textit{Environment \& Background Changing}} \\ 
MLLM-Judge ($\uparrow$) 
& \textcolor{ggg}{5.81}
& \textcolor{ggg}{7.35}
& \textcolor{ggg}{7.37}
&  
& 6.05 
& 2.37  
& \underline{6.60} 
& \textbf{7.18} 
\\ 
Alignment ($\uparrow$) 
& \textcolor{ggg}{28.17}
& \textcolor{ggg}{30.00}
& \textcolor{ggg}{30.04}
&  
& 28.03
& 21.94
& \underline{28.27}
& \textbf{29.15} 
\\ 
Smoothness ($\uparrow$) 
& \textcolor{ggg}{95.76}
& \textcolor{ggg}{96.96}
& \textcolor{ggg}{95.90}
&  
& 89.85
& \underline{95.66}
& \textbf{96.03}
& 94.58 
\\ 
Quality ($\uparrow$) 
& \textcolor{ggg}{61.95}
& \textcolor{ggg}{67.06}
& \textcolor{ggg}{54.58}
&  
& 53.06
& 38.97
& \textbf{54.94}
& \underline{54.25} 
\\ 

\rowcolor[HTML]{d5f5e3} 
\multicolumn{9}{c}{\textit{Visual Feature Changing (Color \& Texture)}} \\ 
MLLM-Judge ($\uparrow$) 
& \textcolor{ggg}{5.86}
& \textcolor{ggg}{6.85}
& \textcolor{ggg}{6.60}
&  
& 6.43 
& 2.14 
& \textbf{7.53} 
& \underline{7.33} 
\\ 
Alignment ($\uparrow$) 
& \textcolor{ggg}{27.99}
& \textcolor{ggg}{29.25}
& \textcolor{ggg}{29.46}
&  
& 27.54
& 23.18
& \textbf{28.88}
& \underline{28.69} 
\\ 
Smoothness ($\uparrow$) 
& \textcolor{ggg}{95.93}
& \textcolor{ggg}{97.10}
& \textcolor{ggg}{95.83}
&  
& 91.71
& \underline{94.75}
& \textbf{96.66}
& 94.52 
\\ 
Quality ($\uparrow$) 
& \textcolor{ggg}{65.80}
& \textcolor{ggg}{69.31}
& \textcolor{ggg}{53.32}
&  
& \underline{58.29}
& 36.27
& \textbf{59.36}
& 57.91 
\\ 

\rowcolor[HTML]{fdebd0} 
\multicolumn{9}{c}{\textit{Stylization}} \\ 
MLLM-Judge ($\uparrow$) 
& \textcolor{ggg}{7.23}
& \textcolor{ggg}{7.62}
& \textcolor{ggg}{8.31}
&  
& 7.41
& 3.71  
& \underline{8.07} 
& \textbf{8.26} 
\\ 
Alignment ($\uparrow$) 
& \textcolor{ggg}{29.84}
& \textcolor{ggg}{30.25}
& \textcolor{ggg}{29.00}
&  
& 27.74
& 22.80
& \underline{29.14}
& \textbf{29.38} 
\\ 
Smoothness ($\uparrow$) 
& \textcolor{ggg}{96.31}
& \textcolor{ggg}{97.23}
& \textcolor{ggg}{96.71}
&  
& 88.97
& 95.62
& \textbf{96.50}
& \underline{95.69} 
\\ 
Quality ($\uparrow$) 
& \textcolor{ggg}{64.05}
& \textcolor{ggg}{68.22}
& \textcolor{ggg}{53.18}
&  
& 54.15
& 35.76
& \textbf{62.59}
& \underline{57.00} 
\\

\rowcolor[HTML]{f6ddcc} 
\multicolumn{9}{c}{\textit{Object Grounding}} \\ 
SSIM ($\uparrow$) 
& \textcolor{ggg}{40.47}
& \textcolor{ggg}{50.46}
& \textcolor{ggg}{47.21}
&  
& 37.98 
& \underline{66.84} 
& 49.65 
& \textbf{70.90} 
\\ 
Jaccard Index $\mathcal{J}$ ($\uparrow$) 
& \textcolor{ggg}{13.85}
& \textcolor{ggg}{19.29}
& \textcolor{ggg}{25.62}
&  
& \underline{19.88} 
& 1.52 
& 13.89 
& \textbf{37.74} 
\\
F-measure $\mathcal{F}$ ($\uparrow$) 
& \textcolor{ggg}{15.50}
& \textcolor{ggg}{16.86}
& \textcolor{ggg}{17.60}
&  
& 12.81 
& 3.07 
& \underline{17.37} 
& \textbf{21.83} 
\\ \midrule

\rowcolor[HTML]{eaeded} 
\multicolumn{9}{c}{\textit{Reasoning Segmentation}} \\ 
SSIM ($\uparrow$) 
& \textcolor{ggg}{-}
& \textcolor{ggg}{-}
& \textcolor{ggg}{-}
&  
& 32.39  
& 44.47 
& \underline{59.86} 
& \textbf{68.41} 
\\ 
Jaccard Index $\mathcal{J}$ ($\uparrow$) 
& \textcolor{ggg}{-}
& \textcolor{ggg}{-}
& \textcolor{ggg}{-}
&  
& 14.02 
& 10.12 
& \underline{16.89} 
& \textbf{22.53}  
\\
F-measure $\mathcal{F}$ ($\uparrow$) 
& \textcolor{ggg}{-}
& \textcolor{ggg}{-}
& \textcolor{ggg}{-}
&  
& 8.07 
& 9.06 
& \underline{10.45} 
& \textbf{15.97}  
\\ \midrule

Avg. Ranking 
& \textcolor{ggg}{2.61} 
& \textcolor{ggg}{1.41}  
& \textcolor{ggg}{1.96} 
&  
& 3.00 
& 3.21
& \underline{2.00} 
& \textbf{1.78} 
\\ 
\bottomrule
\end{tabular}
\endgroup}
\vspace*{-5pt}
\caption{Comparison of video editing task with instructional / non-instructional models on \benchmark. $-$: the task is not capable of non-instructional models. We \textcolor{ggg}{gray} out numbers of non-instructional models that are in different categories. \vspace*{-10pt}}
\label{tab:main}
\end{table*}

%% file: sec/5_conclusion.tex
\section{Conclusion}
We present \model, a unified end-to-end model for instructional video editing that handles diverse pixel-level tasks. \model leverages MLLM for robust instruction understanding and employs a video diffusion model to execute pixel-level edits. 
Our framework uses a single diffusion loss for end-to-end optimization across varied tasks/skills. 
We also introduce a novel synthetic data generation pipeline and \benchmark, a benchmark that assesses a broad range of editing skills. 
Our \model outperforms previous methods as a versatile, all-in-one solution.
We hope our model, data, and benchmark to advance research on instructional generative video models.

%% file: sec/X_suppl.tex
\clearpage
\setcounter{page}{1}
\maketitlesupplementary

\section{Appendix}
\label{sec:appendix}

In this Appedix, we provide extra details on

\begin{itemize}
    \item Implementation details of \model training, and evaluation and baseline evaluations. 
    \item Extra details on our data generation pipeline, including each module's details, prompts for each promptable module, data filtering, and visualization.
    \item Extra visualizations for each task and the comparison with the other 6 strong baseline models.
    \item Limitation and future work discussion.
\end{itemize}

\subsection{Implementation Details}

\noindent\textbf{Model Architecture.} 
Our MLLM is initialized with LLaVA-OneVision-7B (LLaVA-OV)~\cite{li2024llava}. It is a strong MLLM consisting of Qwen2~\cite{yang2024qwen2} LLM with 32K context window, SigLIP~\cite{zhai2023sigmoid} visual encoder, and a 2-layer-MLP projector. LLaVA-OV can handle diverse visual-language tasks (including interleaved-frame, video). It provides a good starting point for our \model to understand complex user instructions and can respond with multiple frame-wise implicit planning thanks to its long context window.
Our video diffusion model is initialized from the instructional image editing model, MagicBrush~\cite{zhang2024magicbrush}. 
We further inflated 2D convolution layers to 3D form and inserted temporal attention layers following AnimateDiff~\cite{guo2023animatediff} to adapt videos.
Our alignment network is a single-layer MLP.
We set 32 grounded task tokens for each frame.

\noindent\textbf{Training Details.}
Our MLLM is initialized with LLaVA-OneVision-7B (LLaVA-OV)~\cite{li2024llava}.
Our VidDM is initialised from the instructional image editing model, MagicBrush~\cite{zhang2024magicbrush} with Stable Diffusion v1.5 backbone~\cite{rombach2022high}. 
We further inflated 2D convolution layers with temporal attention layers, following AnimateDiff~\cite{guo2023animatediff} to adapt videos.
Our \model adopts a 2-stage curriculum training strategy (\cref{sec:training}). 
In the first stage, we fully fine-tune the 2D convolution layers in the UNet, the alignment network, and the task query tokens in the MLLM on image data, with 862M trainable parameters.
In the second stage, we train all 3 dimensions in the UNet, the alignment network, the task query tokens, and a LoRA in the MLLM, leading to 1.3B trainable parameters.
Both stages are trained end-to-end with only a diffusion loss.
More details are in the Appendix.

We keep the VAE encoder and decoder frozen during the entire training process. In the first stage, we keep the MLLM (including visual encoder, MLP projector, and LLM) frozen, and fully fine-tune learnable grounded task queries, alignment network, and diffusion model, leading to around 800M training parameters. 
We set $1e^{-4}$ learning rate, and 96 batch size on each GPU. We use 16 A100 GPUs for the first stage of fine-tuning with 25K steps. 
In the second stage, we insert LoRA~\cite{hu2022lora} modules into the LLM backbone, and inflate diffusion models by inserting extra temporal layers as in AnimateDiff~\cite{guo2023animatediff}. 
We fine-tune LoRA, alignment network, learnable grounded task query tokens, and the diffusion model, leading to around 1.3B trainable parameters.
We set $5e^{-4}$ learning rate, and 1 batch size with 8 gradient accumulation steps on 32 A100 GPUs. 
For LoRA, we set lora rank 64, lora alpha 16, and lora dropout 0.05.
We train the second stage video model 2.5K step with 8 uniformly sampled frames. 

\noindent\textbf{Evaluation and Baseline Details.}
We primarily compare our model with strong instructional editing models~\cite{geng2024instructdiffusion,cheng2023consistent,wu2024towards}. 
Additionally, we include non-instructional editing models~\cite{li2024vidtome,cong2024flatten,geyer2023tokenflow} for completeness, although these are not fair baselines since they are not end-to-end and rely on additional conditions, such as depth maps or intermediate captions.

We randomly sample 3 seeds for both our method and baseline methods.
In our experiments, we use different classifier-free guidance scores ($g_T$ and $g_V$ in~\cref{eq:cfg}) for different skills. Specifically, we set $g_T = 14.5$ and $g_V = 1.5$ for grounding and reasoning segmentation, while for other editing skills, we use $g_T = 10.5$ and $g_V = 2.0$.

For baseline methods, we adopt their default settings (e.g., diffusion steps, guidance scores, frame numbers) as provided in their GitHub repositories. To ensure fair evaluation, we sample the same eight frames from each method's video editing results.

For alignment and smoothness metrics, we use CLIP-B/32 to measure text-image and image-image similarity, averaging across all frames to obtain video-level scores.
For detection metrics, we use GroundingDINO (Swin-T OGC) to detect target objects frame by frame, averaging confidence scores across all frames for the final video-level metric.

For the removal task, where fewer detected objects and lower alignment with the original text prompt are desired, we compute alignment and detection metrics as $1 - \text{value}$.

We compare the model judged best for each video sample. The agreement between human and MLLM judgments is 0.74, whereas the agreement between human and CLIP is only 0.45. We conducted 5 times of the MLLM evaluation and took an average.

\subsection{Data Collection Details}

As mentioned in the earlier~\cref{sec:data}, beyond collecting existing data, we proposed a novel data synthesis pipeline to generate instructional video data by animating images in the instructional image dataset.

Specifically, we first select images from Omni-Edit~\cite{wei2024omniedit}, an instructional image editing dataset with carefully designed tasks/skills.

We first use QWen2-VL~\cite{Qwen2VL} to caption the original image and give an animation prompt to animate the image via CogVideX1.5-I2V~\cite{yang2024cogvideox}. Please refer~\cref{tab:caption_prompt} and~\cref{tab:animation_prompt} to our prompt for caption and animation. After getting the animated video, we utilize AnyV2V~\cite{ku2024anyv2v} to edit the video based on the reference image (edited image from image dataset). The reference image gives a strong prior to maintaining the image dataset's high-quality edit and thus transfer it to the video via the video editing model.

\begin{figure*}
    \centering
    \includegraphics[width=0.7\linewidth]{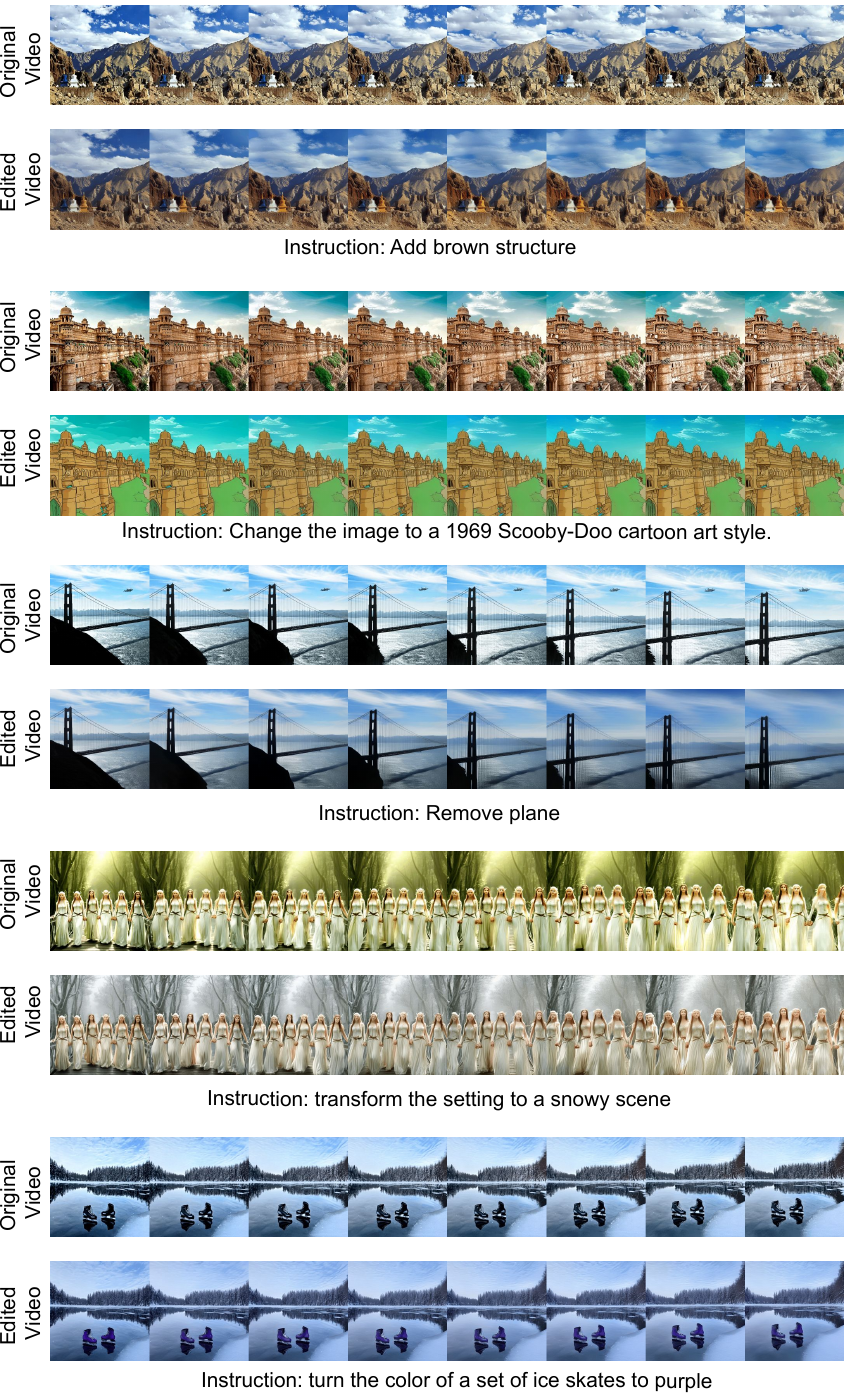}
    \caption{Examples of our generated instructional video editing data.}
    \label{fig:data-vis}
\end{figure*}

Next, we filter out videos by evaluating VBench metrics~\cite{huang2024vbench}, including aesthetic quality, motion smoothness, image quality, subject consistency, and background consistency. We set thresholds at 0.6 for aesthetic quality, 65 for imaging quality, 0.9 for motion smoothness, subject consistency, and background consistency. We provide our generated data visualization in~\cref{fig:data-vis}.

\subsection{More Quantative Results \& Discussion}

\input{materials/grounding}

\noindent\textbf{Video Concept Grounding \& Reasoning Segmentation}
We include additional results on video concept grounding and reasoning segmentation in \cref{tab:grounding}. \model outperforms the diffusion-based baseline by a significant margin, showcasing its superior ability to accurately locate fine-grained object references and handle complex reasoning tasks.
We hypothesize that through grounded generation, \model demonstrates remarkable precision in concept editing. For example, as shown in Fig.~\ref{fig:app_remove} in the Appendix, \model can remove the woman without altering the nearby girl.

\subsection{Limitation and Future Works}
Our current method, \model, is built upon Stable-Diffusion 1.5, which inevitably constrains its editing quality compared to cutting-edge video generation models that rely on DiT or flow-based architectures. 
In addition, the video outputs we produce are relatively short, lagging behind some recent state-of-the-art methods in terms of length and temporal consistency. 
Furthermore, we observe increased editing artifacts when incorporating large amounts of grounding data, suggesting that multi-task data mixture strategies play a key role in maintaining high-quality edits.

Despite these limitations, our results demonstrate promising directions for improvement in terms of model design, data curation, and evaluation. 
Future work could explore integrating more advanced base architectures (e.g., DiT~\cite{yang2024cogvideox, kong2024hunyuanvideo} or flow-based models), extending the maximum video duration, developing more systematic data~\cite{hu2024vivid} with more advanced method~\cite{liu2024generative} and carefully designed mixture strategies to balance fidelity and flexibility, and conducting scalable training. 
We hope our findings will inspire further research into these directions, pushing the boundaries of instructional video editing performance.
\begin{figure}
    \centering
    \includegraphics[width=0.85\linewidth]{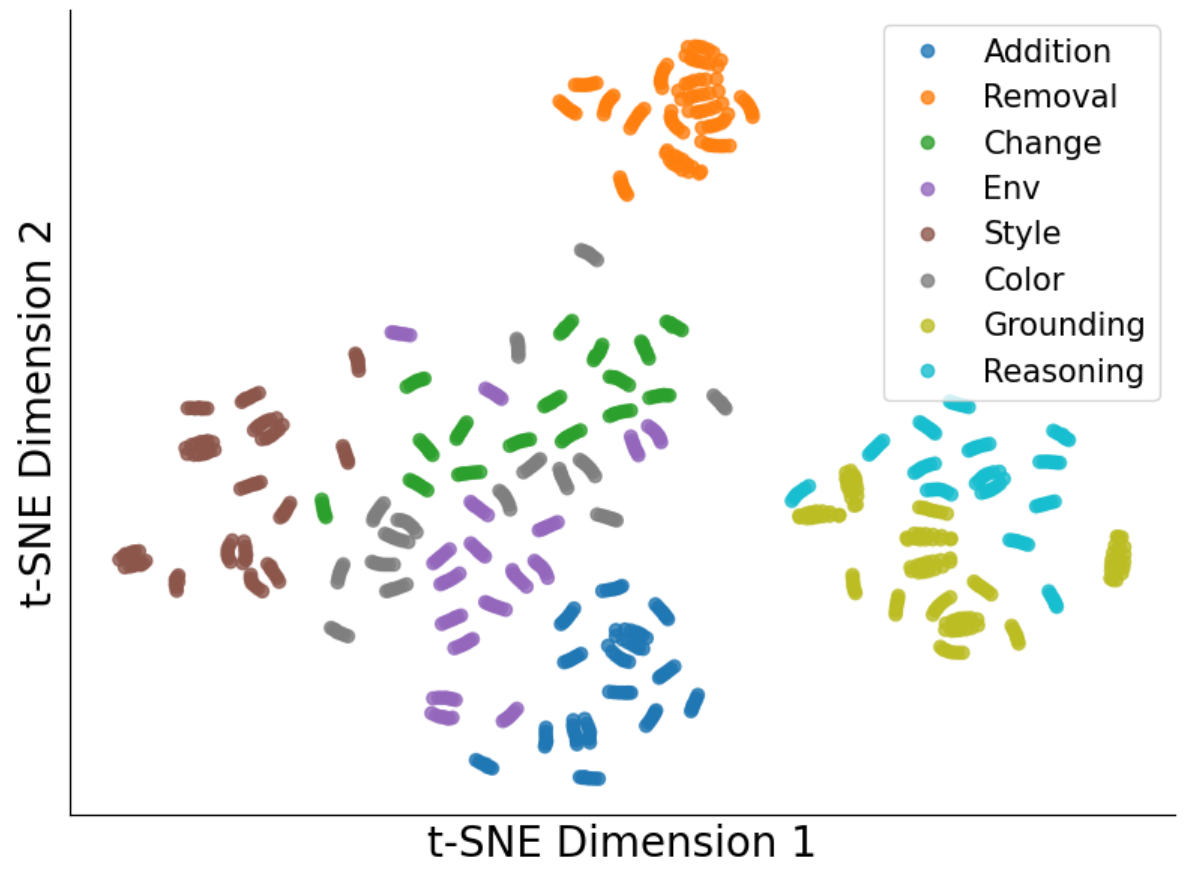}
    \caption{t-SNE Visualization of different task query distribution. Different colors represent different tasks/skills. Best view in color.}
    \label{fig:tsne}
\end{figure}

\noindent\textbf{Task Query Visualization \& Analysis via t-SNE.}
To analyze task/skill correlations, we project their grounded queries into lower-dimensional spaces using PCA and t-SNE. As shown in~\cref{fig:tsne}, distinct clusters form for each category (e.g., Addition), indicating effective differentiation by the model.
\textit{Reasoning} and \textit{Grounding} appear together on the right. It may be because they both require cognitive/semantic understanding or logical reference.
\textit{Color}, \textit{Env}, and \textit{Change} clusters are closer to each other, indicating that the model views them as similar operations focusing on changing different visual attributes. 
\textit{Style} lies in the lower-left region but remains relatively close to \textit{Color}, \textit{Env}, and \textit{Change}. 
This proximity may reflect that ``stylization'' is conceptually similar to these visual attribute tasks, although it targets different transformations.
\textit{Removal} stands apart on the top, especially distant from \textit{Addition}, indicating the model perceives them as distinct rather than inverse operations. In contrast, \textit{Addition} lies closer to tasks like \textit{Reasoning} and \textit{Grounding}. It suggests that the act of adding elements may rely on similar semantic or referential processes (e.g., deciding what to add and how to reference the newly added element).

\begin{table*}[t!]
\centering
\begin{minipage}{1\columnwidth}\vspace{20mm}    
    \centering
    \RawFloats
    \caption{Qwen2-VL prompt for Image caption.}
    \begin{tcolorbox} 
    
        \centering
        \hspace{-6mm}
        \begin{tabular}{p{0.99\columnwidth}}
        \hspace{1mm}
        \begin{minipage}{0.99\columnwidth}
Please describe this image shortly, try to capture main details in the image. \\
Here are some examples of image caption styles:\\

1. A Couple In A Public Display Of Affection\\
2. A kitten turning its head on a wooden floor\\
3. An Old Man Doing Exercises For The Body And Mind\\
4. Man Walking\\

Now, please describe the given image briefly in one sentence, please do not say something like 'The image shows...' or 'The image depicts...'.

        \end{minipage}
        \end{tabular}
    \end{tcolorbox}
\label{tab:caption_prompt}
\end{minipage}
\end{table*}

\begin{table*}[t!]
\centering
\begin{minipage}{1\columnwidth}\vspace{20mm}    
    \centering
    \RawFloats
    \caption{Qwen2-VL prompt for generating animation prompt.}
    \begin{tcolorbox} 
    
        \centering
        \hspace{-6mm}
        \begin{tabular}{p{0.99\columnwidth}}
        \hspace{1mm}
        \begin{minipage}{0.99\columnwidth}
I want to animate this image using an Image-Text-to-Video model. Your task is to generate a detailed and reasonable text prompt that describes how the image should be animated.\\

Guidelines:\\

1. Clarity \& Realism - The animation description should be logical based on the given image, ensuring the movement makes sense for the scene.\\

2. Short \& Vivid Description - Use expressive language to guide the animation model effectively, ensuring high-quality and visually engaging results.\\

Ensure that your animation prompt aligns with the content of the provided image and describes a visually compelling motion sequence.\\

Do not output animation prompts that contain objects/scenes not included in the given image.\\

Make sure the prompt is short in 1-2 sentences.

        \end{minipage}
        \end{tabular}
    \end{tcolorbox}
\label{tab:animation_prompt}
\end{minipage}
\end{table*}

\begin{table*}[t!]
\centering
\begin{minipage}{1\columnwidth}\vspace{20mm}    
    \centering
    \RawFloats
    \caption{GPT-4o prompt for MLLM-as-a-Judge for automatic instructional video editing evaluation.}
    \begin{tcolorbox} 
    
        \centering
        \hspace{-6mm}
        \begin{tabular}{p{0.99\columnwidth}}
        \hspace{1mm}
        \begin{minipage}{0.99\columnwidth}
        \textbf{User} \\
    You are an evaluator for instructional video editing tasks. Your job is to assess how well the edited video fulfills the user's specific instructions. \\
    I will provide: \\
    1. The original video (first GIF) \\
    2. The edited video (second GIF) \\
    3. The user's instruction: [user instruction] \\
    Please evaluate the editing result using the following format:\\
    INSTRUCTION: [Repeat the user's instruction]\\
    EVALUATION:\\
    - Accuracy score (1-10): \textcolor{blue}{[Your score]}\\
    - Quality score (1-10): \textcolor{blue}{[Your score]} \\
    - Appropriateness score (1-10): \textcolor{blue}{[Your score]} \\
    - Overall score (1-10): \textcolor{blue}{[Your final score]} \\
    \\
    EXPLANATION: \textcolor{blue}{[Provide a brief justification for your scores, highlighting specific strengths and weaknesses of the edit]}\\
    RECOMMENDATION: \textcolor{blue}{[Optional suggestions for improvement]}\\

    When scoring, consider:\\
    - Accuracy: Does the edit precisely follow the given instruction?
    - Quality: Is the edit visually seamless and natural-looking?
    - Appropriateness: Does the edit maintain coherence with the original video context?\\

    The overall scale is:\\
    1-3: Poor - Major issues with the edit\\
    4-6: Acceptable - Follows instruction but with noticeable flaws\\
    7-8: Good - Clear, effective edit with minor issues\\
    9-10: Excellent - Flawless execution of the instruction\\
        \rule[0.25\baselineskip]{\textwidth}{1pt}
        \textbf{Assistant} \\
        {\textcolor{blue}{Scores}}, {\textcolor{blue}{Explanation}}, {\textcolor{blue}{Recommendation}}
        \end{minipage}
        \end{tabular}
    \end{tcolorbox}
\label{tab:evaluation_prompt}
\end{minipage}
\end{table*}

\subsection{Extra Visualization}

\begin{figure*}
    \centering
    \includegraphics[width=\linewidth]{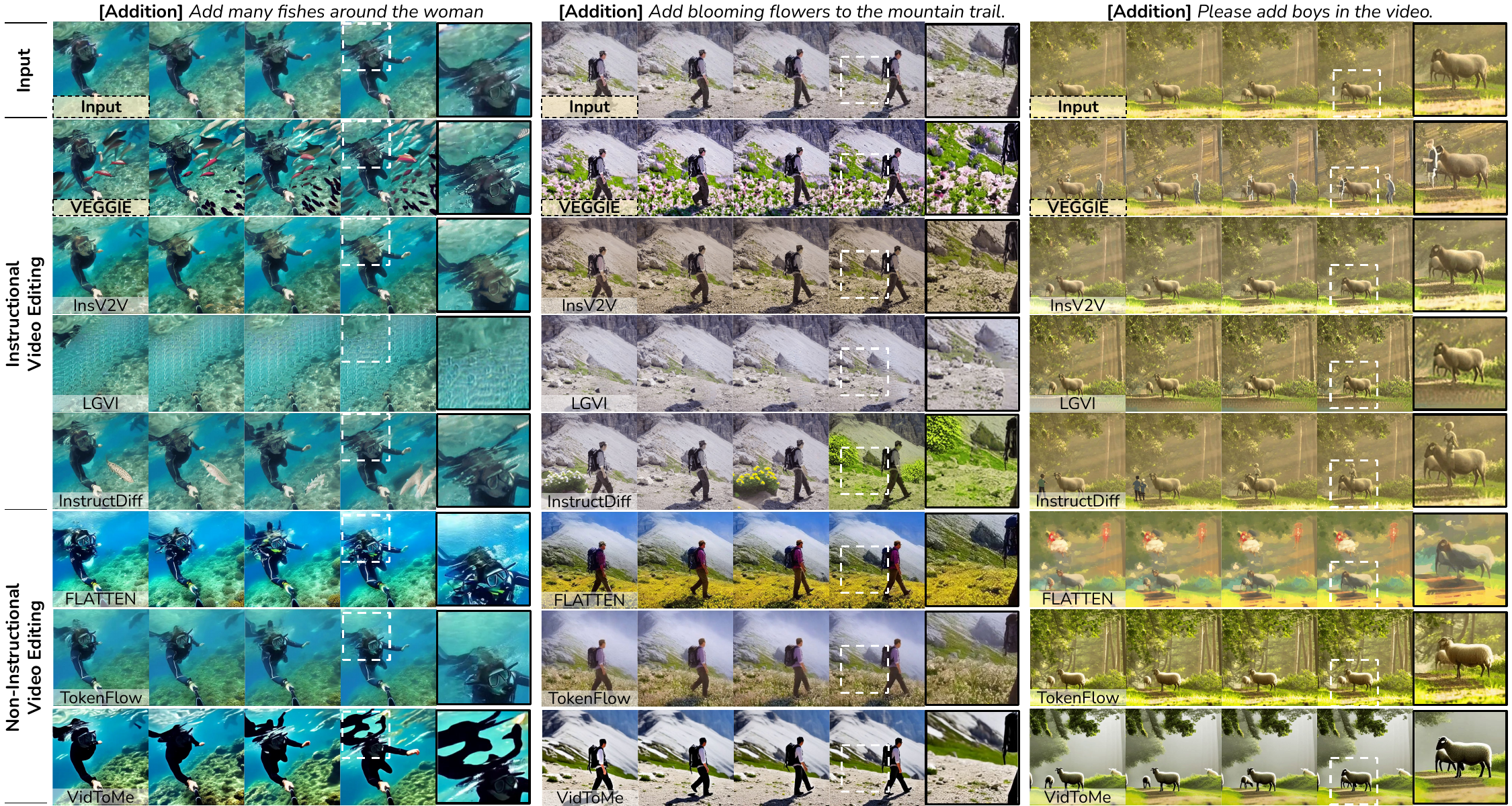}
    \caption{More Examples of Concept Addition.}
    \label{fig:app_add}
\end{figure*}

\begin{figure*}
    \centering
    \includegraphics[width=\linewidth]{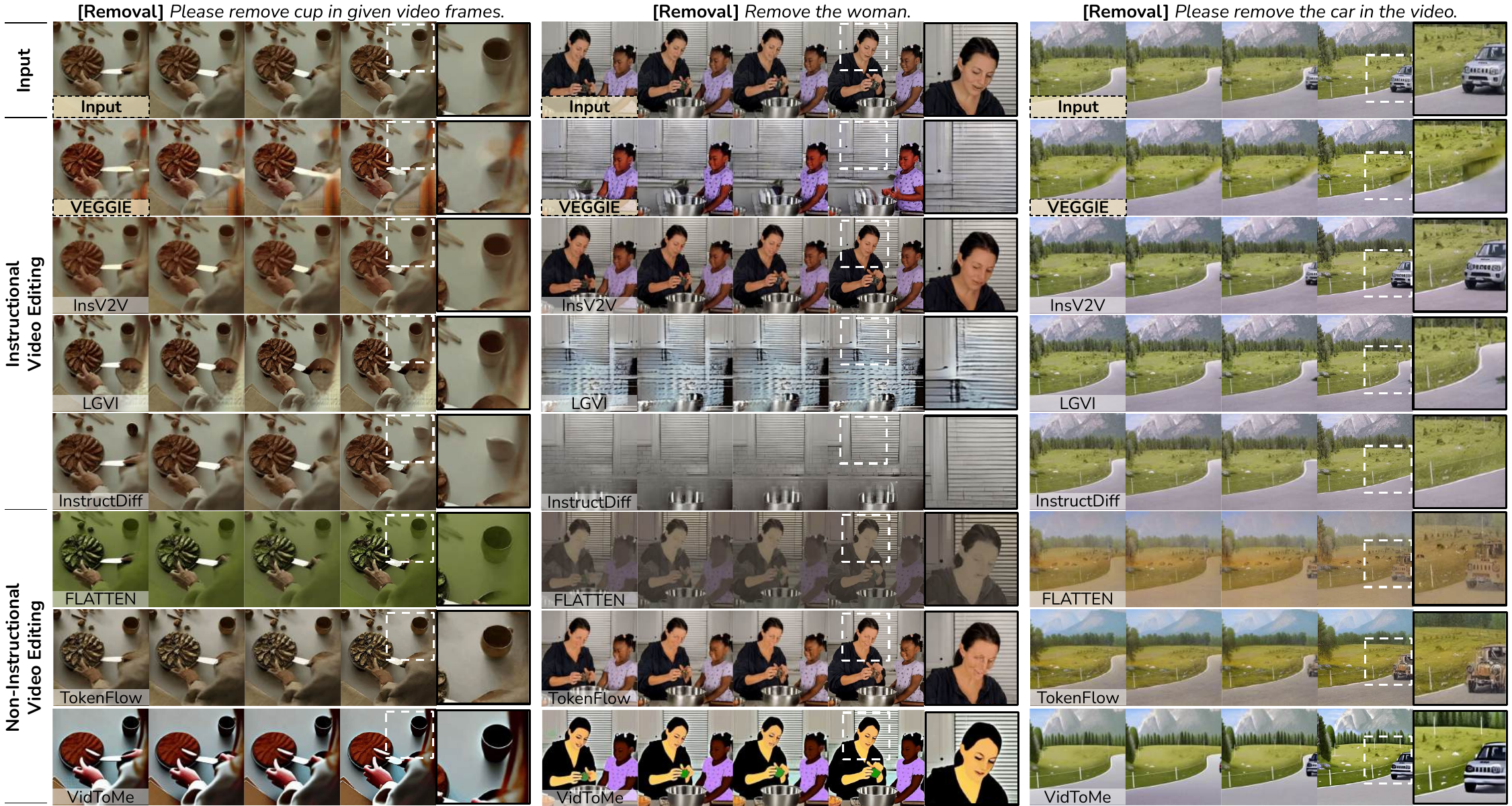}
    \caption{More Examples of Concept Removal.}
    \label{fig:app_remove}
\end{figure*}

\begin{figure*}
    \centering
    \includegraphics[width=\linewidth]{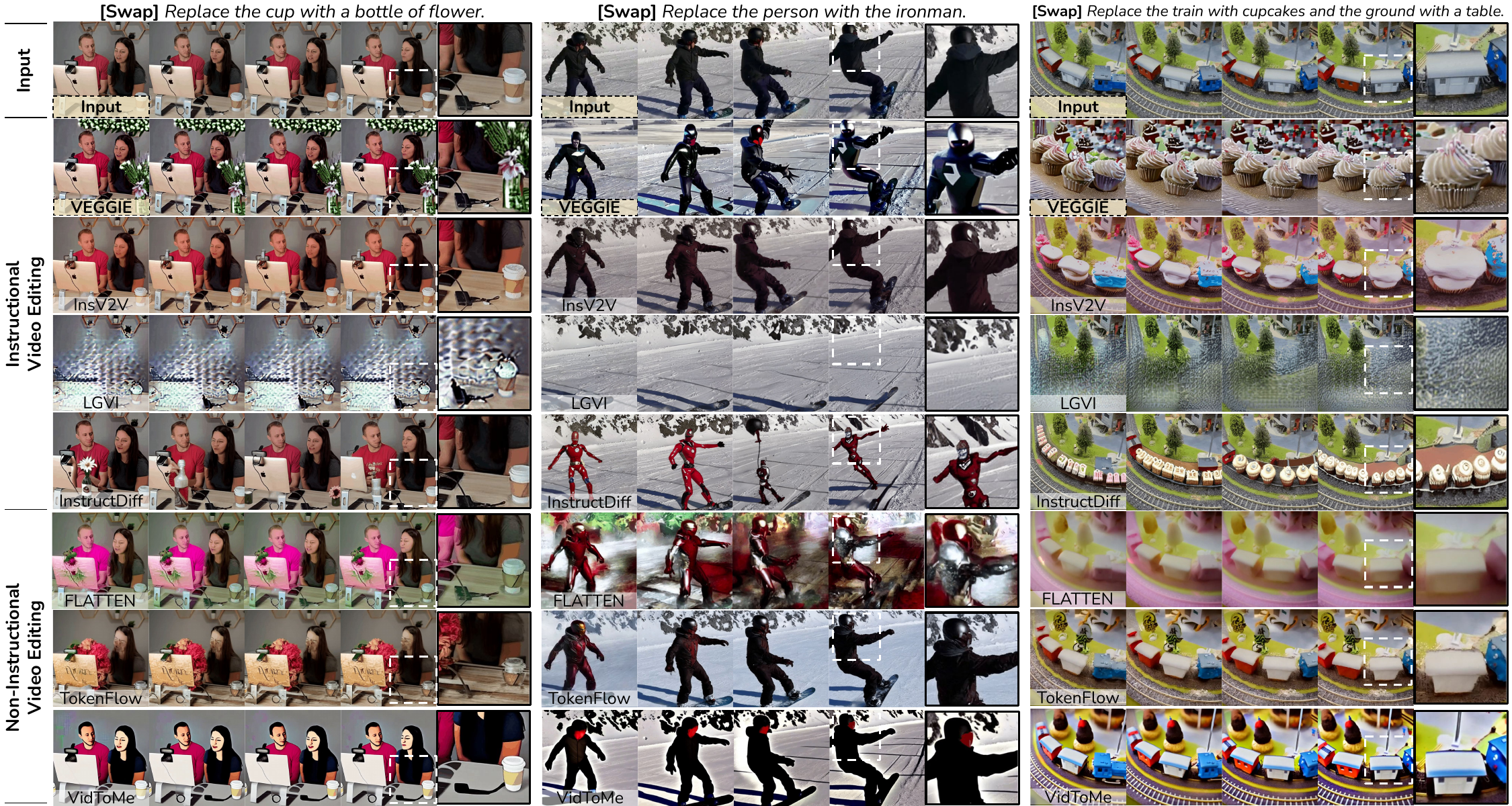}
    \caption{More Examples of Object Changes.}
    \label{fig:app_swap}
\end{figure*}

\begin{figure*}
    \centering
    \includegraphics[width=\linewidth]{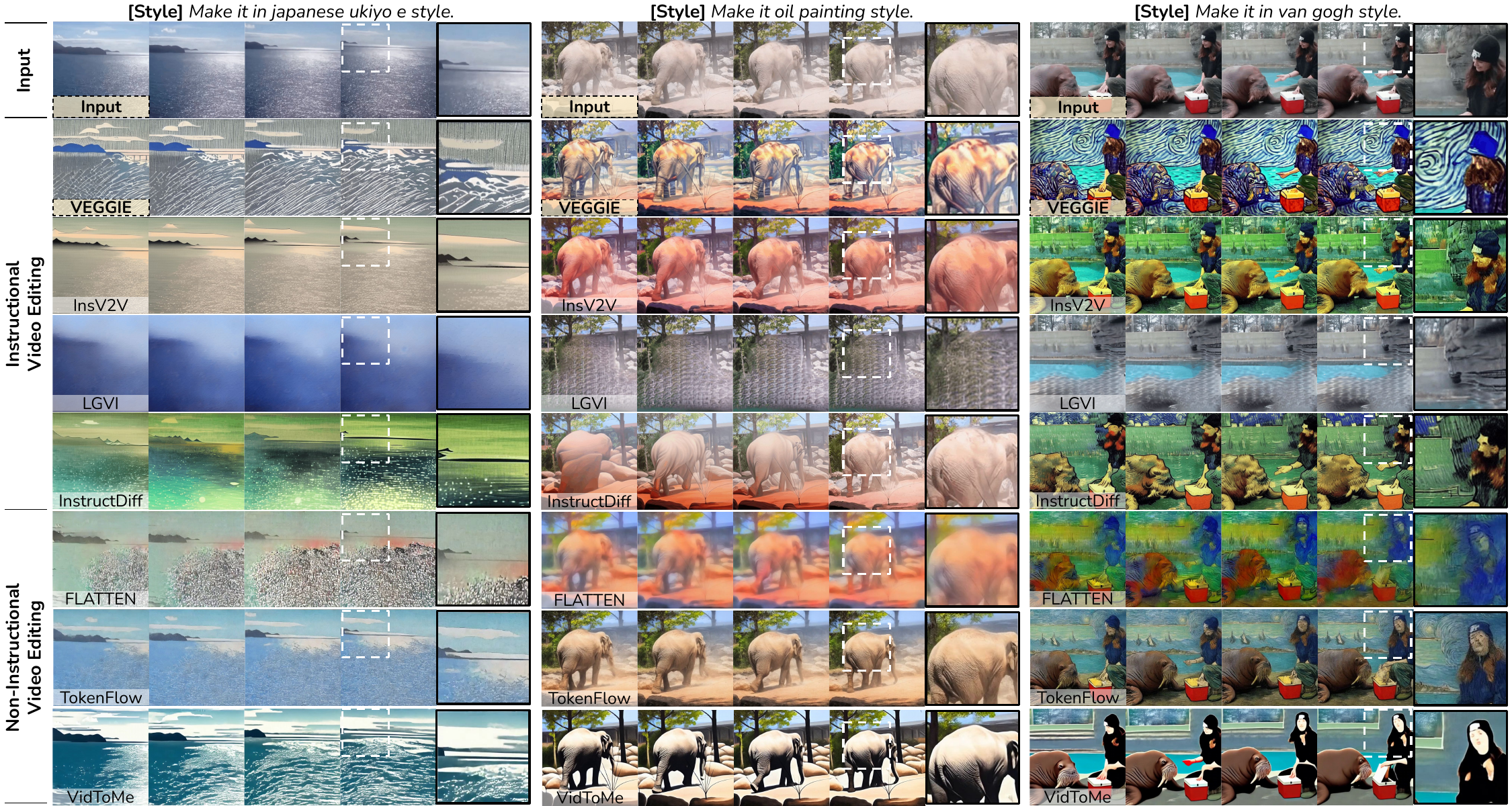}
    \caption{More Examples of Stylization.}
    \label{fig:app_style}
\end{figure*}

\begin{figure*}
    \centering
    \includegraphics[width=\linewidth]{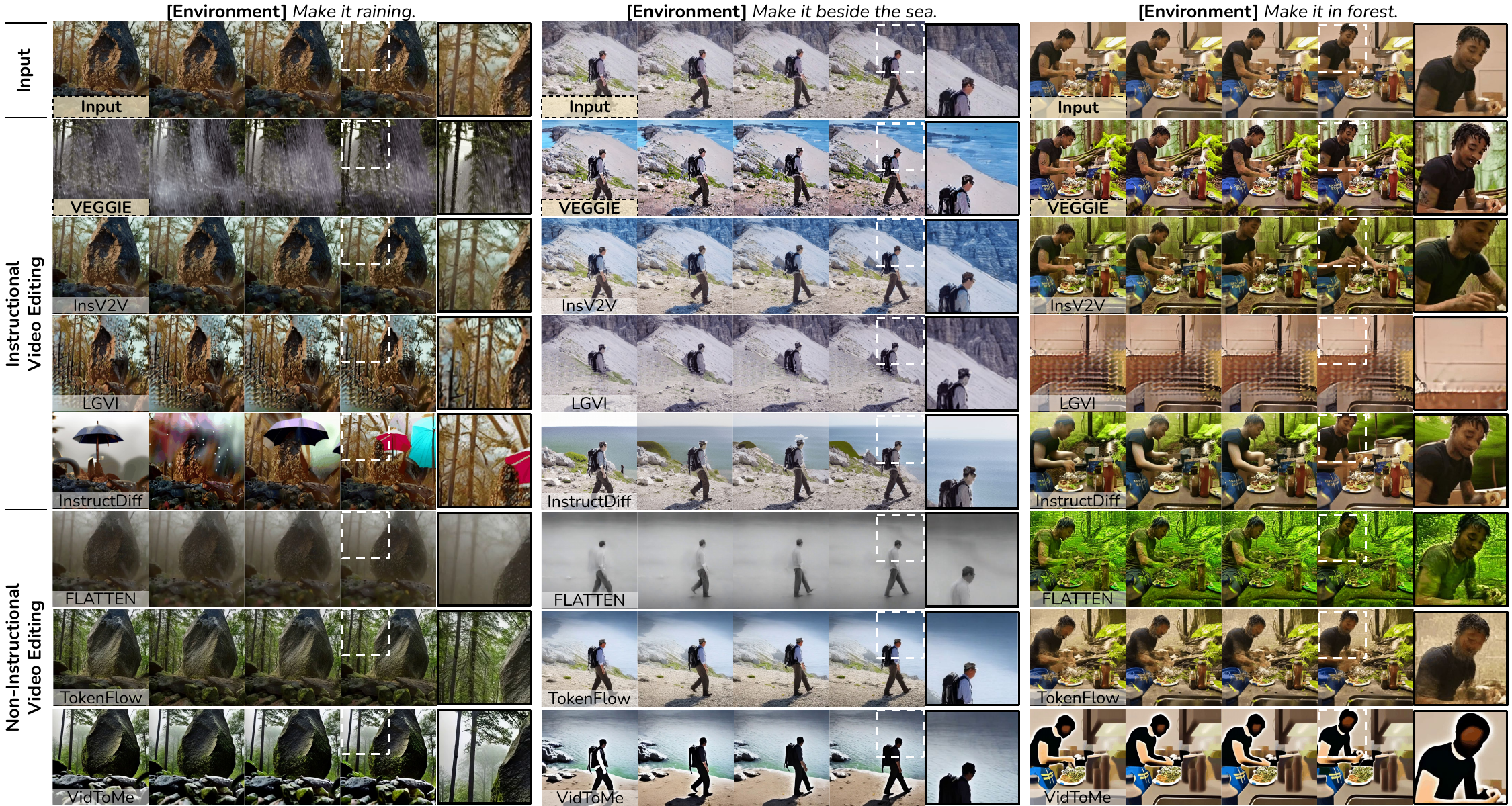}
    \caption{More Examples of Environment and Background Editing.}
    \label{fig:app_env}
\end{figure*}

\begin{figure*}
    \centering
    \includegraphics[width=\linewidth]{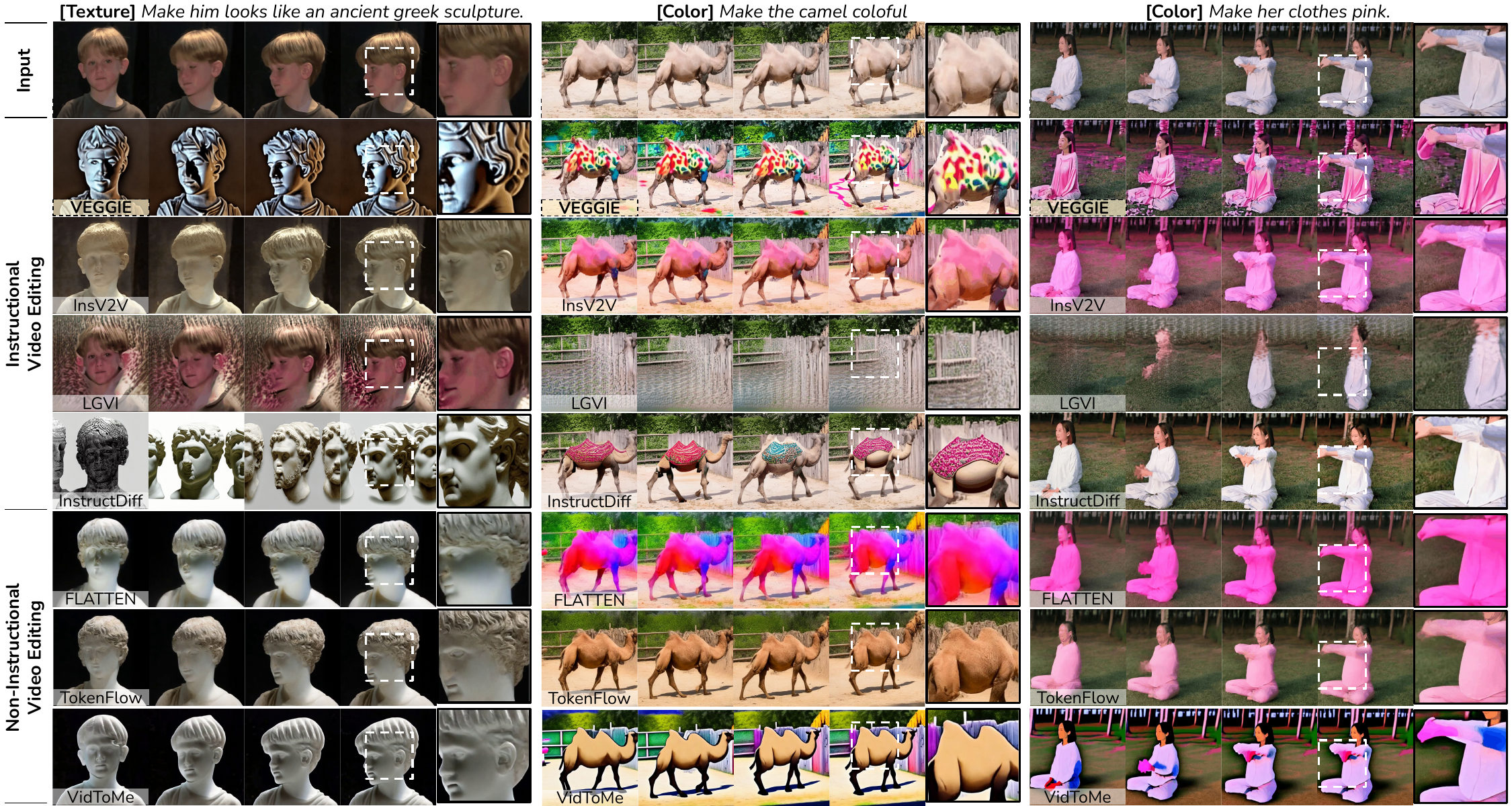}
    \caption{More Examples of Visual Features Editing.}
    \label{fig:app_feature}
\end{figure*}

\begin{figure*}
    \centering
    \includegraphics[width=\linewidth]{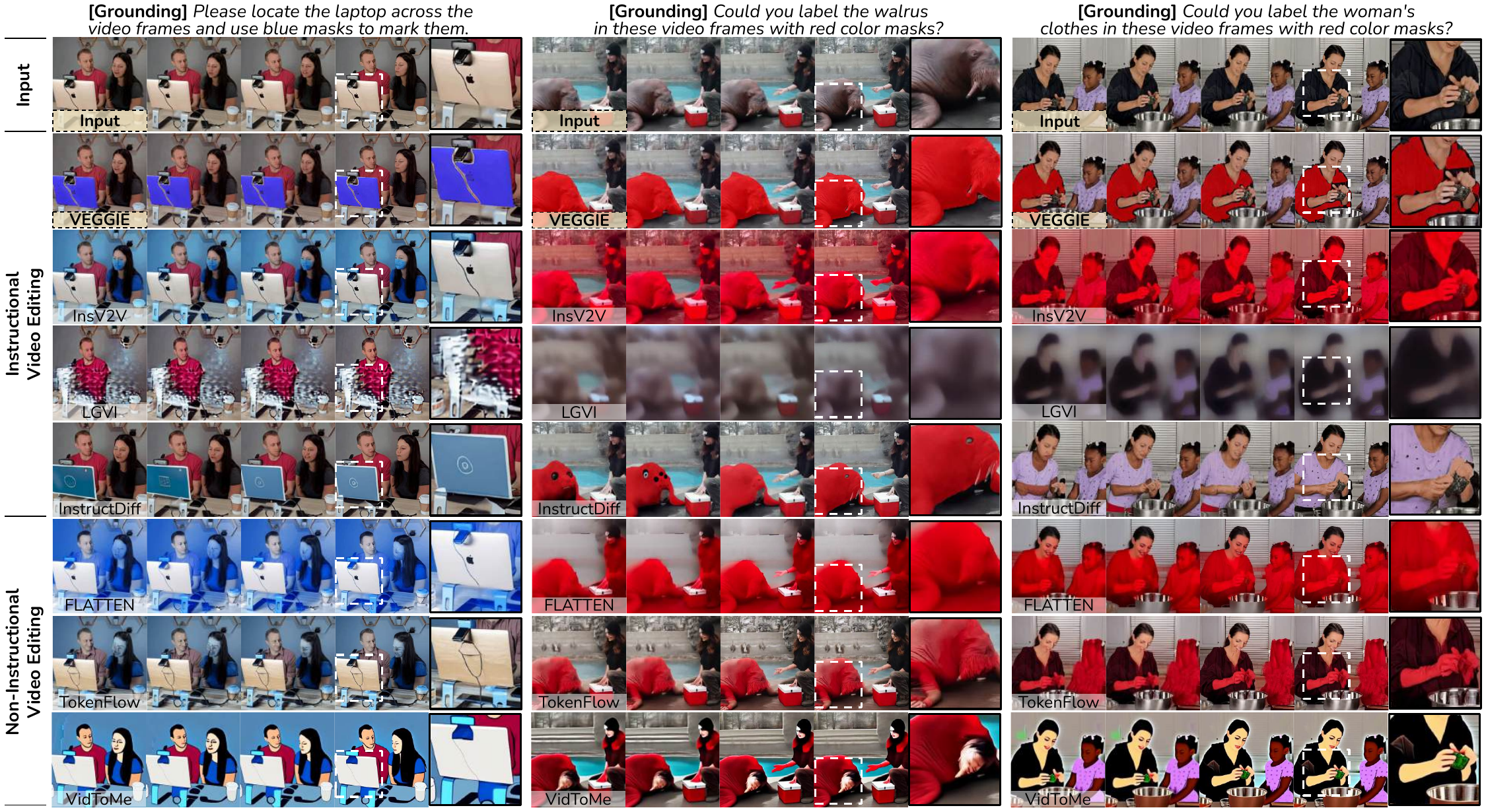}
    \caption{More Examples of Object Grounding.}
    \label{fig:app_ground}
\end{figure*}

We provide extra visualization in~\cref{fig:app_remove,fig:app_add,fig:app_swap,fig:app_style,fig:app_env,fig:app_feature,fig:app_ground}

\begin{figure*}
    \centering
    \includegraphics[width=\linewidth]{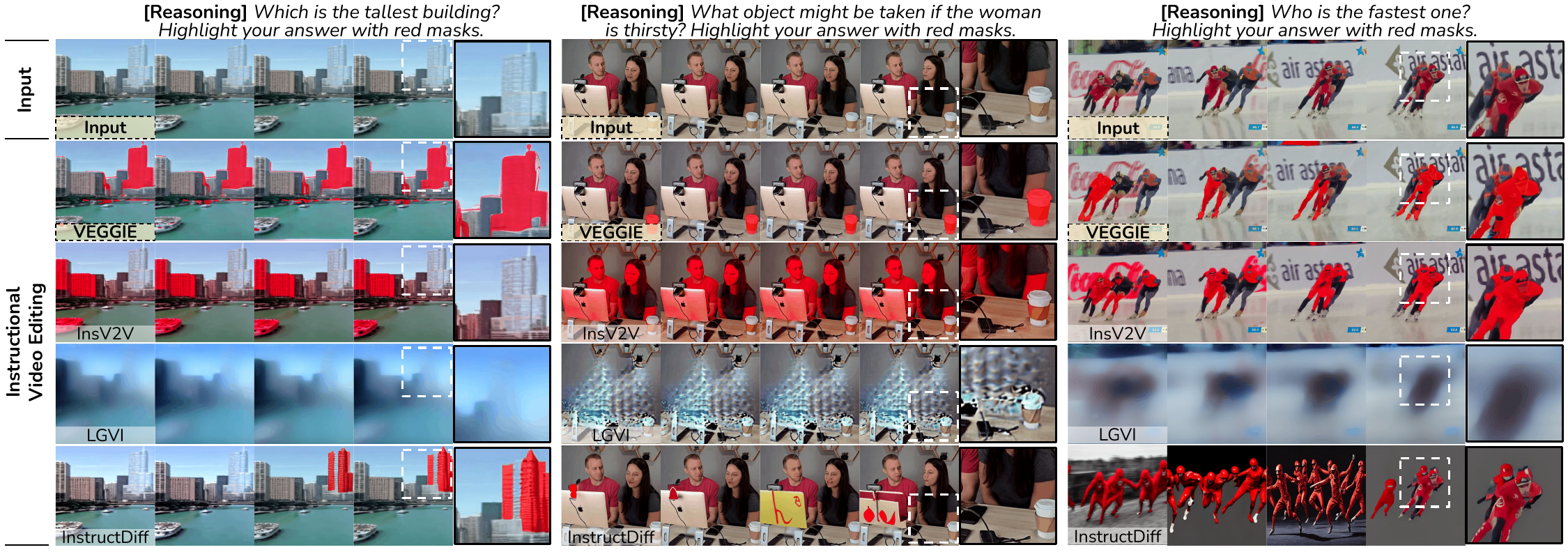}
    \caption{More Examples of Object Reasoning Segmentation.}
    \label{fig:app_reasoning}
\end{figure*}

%% file: materials/grounding.tex
\begin{table}[]
\centering
\scalebox{0.9}{
\begingroup
\setlength{\tabcolsep}{4pt}
\renewcommand{\arraystretch}{0.9}
\hspace{-10pt}
\begin{tabular}{lcccccc}
\toprule
\multirow{2}{*}{\textbf{Methods}} & \multicolumn{3}{c}{\textbf{Grounding}}                                                              & \multicolumn{3}{c}{\textbf{Reasoning}} \\ \cmidrule(lr){2-7} 
                                  & \multicolumn{1}{c}{$\mathcal{J}$} & \multicolumn{1}{c}{$\mathcal{F}$} & \multicolumn{1}{c}{$\mathcal{J\&F}$} & \multicolumn{1}{c}{$\mathcal{J}$} & \multicolumn{1}{c}{$\mathcal{F}$} & \multicolumn{1}{c}{$\mathcal{J\&F}$} \\ \midrule
\rowcolor[HTML]{eaeded} \multicolumn{7}{c}{\textit{Segmentation Models}} \\ 
HTR~\cite{miao2024htr} &\textcolor{ggg}{47.11}&\textcolor{ggg}{47.60}&\textcolor{ggg}{47.35}&\textcolor{ggg}{20.01}&\textcolor{ggg}{28.02}&\textcolor{ggg}{24.01}\\
VideoLISA~\cite{bai2024one} &\textcolor{ggg}{53.23}&\textcolor{ggg}{54.37}&\textcolor{ggg}{53.80}&\textcolor{ggg}{38.48}&\textcolor{ggg}{39.20}&\textcolor{ggg}{38.84}\\
MoRA~\cite{deng2024groundmore} &\textcolor{ggg}{57.73}&\textcolor{ggg}{53.63}&\textcolor{ggg}{55.68}&\textcolor{ggg}{38.92}&\textcolor{ggg}{37.48}&\textcolor{ggg}{40.36}\\
\rowcolor[HTML]{eaeded} \multicolumn{7}{c}{\textit{Generative Editing Models}} \\
InstructDiff~\cite{geng2024instructdiffusion} & 19.88 & 12.81 & 16.35 & 14.02 &8.07 & 11.05 \\
InsV2V~\cite{cheng2023consistent} & 13.89 & 17.37 & 15.63 & 16.89 & 10.45 & 13.67 \\
\textbf{\model} (Ours) & \textbf{37.74} & \textbf{21.83} & \textbf{29.79}& \textbf{22.53}& \textbf{15.97} & \textbf{19.25}\\ \bottomrule
\end{tabular}
\endgroup}
\caption{Comparison of video concept grounding and reasoning segmentation tasks with other instructional generative models and expert segmentation models.}
\label{tab:grounding}
\end{table}